\documentclass{article}

\usepackage[preprint]{neurips_2023}

\usepackage[utf8]{inputenc} 
\usepackage[T1]{fontenc}    
\usepackage{hyperref}       
\usepackage{url}            
\usepackage{booktabs}       
\usepackage{amsfonts}       
\usepackage{nicefrac}       
\usepackage{amsmath}
\usepackage{adjustbox}
\usepackage{microtype}      
\usepackage{xcolor}         
\usepackage{graphicx}
\usepackage{marvosym}
\usepackage{amssymb} 
\usepackage{wrapfig} 
\usepackage{multirow} 
\newtheorem{proposition}{Proposition} 

\title{Machine Learning Insides OptVerse AI Solver: \\Design Principles and Applications}

\author{
\textbf{Xijun~Li}\textsuperscript{1\Letter}\,\,,\,
\textbf{Fangzhou~Zhu}\textsuperscript{1},\,
\textbf{Hui-Ling~Zhen}\textsuperscript{1},\,
\textbf{Weilin~Luo}\textsuperscript{1},\,
\textbf{Meng~Lu}\textsuperscript{1},\,\\
\textbf{Yimin~Huang}\textsuperscript{1},\,
\textbf{Zhenan~Fan}\textsuperscript{3},\,
\textbf{Zirui~Zhou}\textsuperscript{3},\,
\textbf{Yufei~Kuang}\textsuperscript{4},\,
\textbf{Zhihai~Wang}\textsuperscript{4},\,\\
\textbf{Zijie~Geng}\textsuperscript{4},\,
\textbf{Yang~Li}\textsuperscript{5},\,
\textbf{Haoyang~Liu}\textsuperscript{4},\,
\textbf{Zhiwu~An}\textsuperscript{2},\,
\textbf{Muming~Yang}\textsuperscript{2},\,\\
\textbf{Jianshu~Li}\textsuperscript{2},\,
\textbf{Jie~Wang}\textsuperscript{4},\,
\textbf{Junchi~Yan}\textsuperscript{5},\,
\textbf{Defeng~Sun}\textsuperscript{6},\,
\textbf{Tao~Zhong}\textsuperscript{1},\,\\
\textbf{Yong~Zhang}\textsuperscript{3},\,
\textbf{Jia~Zeng}\textsuperscript{1},\,
\textbf{Mingxuan~Yuan}\textsuperscript{1},\,
\textbf{Jianye~Hao}\textsuperscript{1},\,
\textbf{Jun~Yao}\textsuperscript{1},\,
\textbf{Kun~Mao}\textsuperscript{2\Letter}\\
\textsuperscript{1}Huawei Noah’s Ark Lab\thanks{Correspondence to: Xijun Li <xijun.li@huawei.com> and Kun Mao <maokun@huawei.com>}\\\textsuperscript{2}Huawei Cloud EI Service Product Dept.\thanks{This work was jointly developed by Huawei 2012 Labs and Huawei Cloud. We also thank the work from Huawei Vancouver Research Center, Huawei Moscow Research Center, Huawei Minsk Research Center, and Huawei Munich Research Center.}\\\textsuperscript{3}Huawei Vancouver Research Center\\
\textsuperscript{4}University of Science and Technology of China\\
\textsuperscript{5}Shanghai Jiao Tong University\\
\textsuperscript{6}Hong Kong Polytechnic University\\
}
\begin{document}

\maketitle

\begin{abstract}
In an era of digital ubiquity, efficient resource management and decision-making are paramount across numerous industries. To this end, we present a comprehensive study on the integration of machine learning (ML) techniques into Huawei Cloud's OptVerse AI Solver, which aims to mitigate the scarcity of real-world mathematical programming instances, and to surpass the capabilities of traditional optimization techniques. We showcase our methods for generating complex SAT and MILP instances utilizing generative models that mirror multifaceted structures of real-world problem. Furthermore, we introduce a training framework leveraging augmentation policies to maintain solvers’ utility in dynamic environments. Besides the data generation and augmentation, our proposed approaches also include novel ML-driven policies for personalized solver strategies, with an emphasis on applications like graph convolutional networks for initial basis selection and reinforcement learning for advanced presolving and cut selection. Additionally, we detail the incorporation of state-of-the-art parameter tuning algorithms which markedly elevate solver performance. Compared with traditional solvers such as Cplex and SCIP, our ML-augmented OptVerse AI Solver demonstrates superior speed and precision across both established benchmarks and real-world scenarios, reinforcing the practical imperative and effectiveness of machine learning techniques in mathematical programming solvers.
\end{abstract}

\section{Introduction}

Digital construction is one of the most pivotal tasks of thousands of trades in this era. Guided by this objective, the enhancement of management efficiency across industries, digital decision-making, and the improved utilization of resources stand as obligatory challenges to be addressed. Huawei Cloud's OptVerse AI Solver is not solely applicable to the port industry~\cite{tianjin_port} but extends its utility across a myriad of sectors. For instances, 'Black Friday' shopping, the question arises as to how one can manage complex storage and logistics; in the face of a surge in orders, how can one manage tens of thousands of employees and hundreds of factories to achieve the maximal utilization of resources? At major airports, how can one ensure that thousands of daily flights maximize the use of jet bridges? For practitioners in manufacturing, retail, logistics, and other sectors, such problems are undoubtedly familiar. They are required to make similar decisions daily, but how can one achieve optimal resource allocation? The answers to these questions can be found with the assistance of the OptVerse AI Solver. 

In the pursuit of optimizing the performance of OptVerse AI solver, the integration of machine learning (ML) techniques emerges as an imperative strategy. The necessity of embracing ML within our solvers is primarily driven by the quest to address the evolving complexities and diversities inherent in real-world mathematical programming problems. As these real-world mathematical programming instances are often scarce and encumbered by data curation challenges and proprietary restrictions, the induction of machine learning not only compensates for this scarcity but also fosters significant enhancements in solver capabilities.

Therefore, ML-driven data generation and augmentation techniques play a crucial role in the development and fine-tuning of mathematical programming solvers. Generating novel mathematical programming instances artificially extends the horizons of training and evaluation environments, thus contributing to the robustness and discovery of solver algorithms. For instance, our proposed HardSATGEN and G2MILP leverage generative models to create sophisticated and strategically challenging SAT and MILP instances respectively, which mirror real-world problem structures. On the other hand, data augmentation aims to enhance solver generalizability, allowing them to adapt to shifts in environments and out-of-distribution instances within constrained data availability scenarios. By using our proposed AdaSolver training framework, OptVerse AI solver is equipped with augmentation policies that are both computationally efficient and effective in modifying existing instances to suit the dynamic problem landscapes.

Moreover, the infusion of machine learning into policy learning for OptVerse AI solver has revolutionized decision-making processes. ML-driven policies enable solvers to personalize strategies for individual problem instances, dramatically increasing convergence rates and improving solver performance. Our specialized applications, such as Graph Convolutional Networks (GCNs) for Initial Basis Selection, exemplify the ability of ML to exploit patterns in problem instances. Similarly, techniques like reinforcement learning for presolve operations and hierarchically-structured sequence models for Cut Selection reflect the synergetic integration of policy learning with solver optimization. These innovations, including the Neural Diving heuristic with its GCNN-based approach to curation of binary variable assignments, hallmark a pivotal transition towards computationally adept and scalable solvers.

Besides, we all know that the advanced open-source/commercial solver~\cite{scip, gurobi_solver, cplex}, etc are equipped with hundreds of parameters to be tuned, due to the complexity of features. These parameters have huge impacts on the performance of solvers.
Machine learning is instrumental in parameter tuning for the solver's hyper-parameter space, thereby improving search efficiency and solution quality through a data-informed, systematic trial-and-error process. Machine learning algorithms such as HEBO~\cite{hebo} and Transformber BO~\cite{alex2023end} are vital for orchestrating trial execution and strategically managing computational resources across hardware configurations.

The significance and necessity of integrating machine learning techniques into the OptVerse AI solver are twofold: (1) they expand the pool of mathematical programming instances available for solver refinement and performance evaluation, and (2) they endow solvers with the cognitive flexibility to personalize and adapt strategies in real time. This potent amalgamation of machine learning and operational research propagates an era of computational ingenuity, leading to solvers that are not only more efficient but also inherently equipped to tackle the intricacies of modern mathematical programming challenges. Our proposed methods and developed tools have already helped improve the performance, in terms of speed and accuracy, of our OptVerse AI solver by a large margin over both real-life problem and well-recognized mathematical programming solver benchmark.

Compared to traditional mathematical programming solver such as Gurobi, Cplex and SCIP, etc, we actively bring machine learning techniques into our OptVerse AI Solver, aiming to optimize the solver to the extreme. In this manuscript, we first simply provide a reflection on the trend of integrating machine learning techniques into mathematical programming solvers. Then we state our design principles for integrating machine learning techniques in the OptVerse AI solver and corresponding applications and its implementation. Finally, the manuscript concludes with our outlook.
\section{Related work}
\begin{figure}[t]
    \centering
    \includegraphics[width=\linewidth]{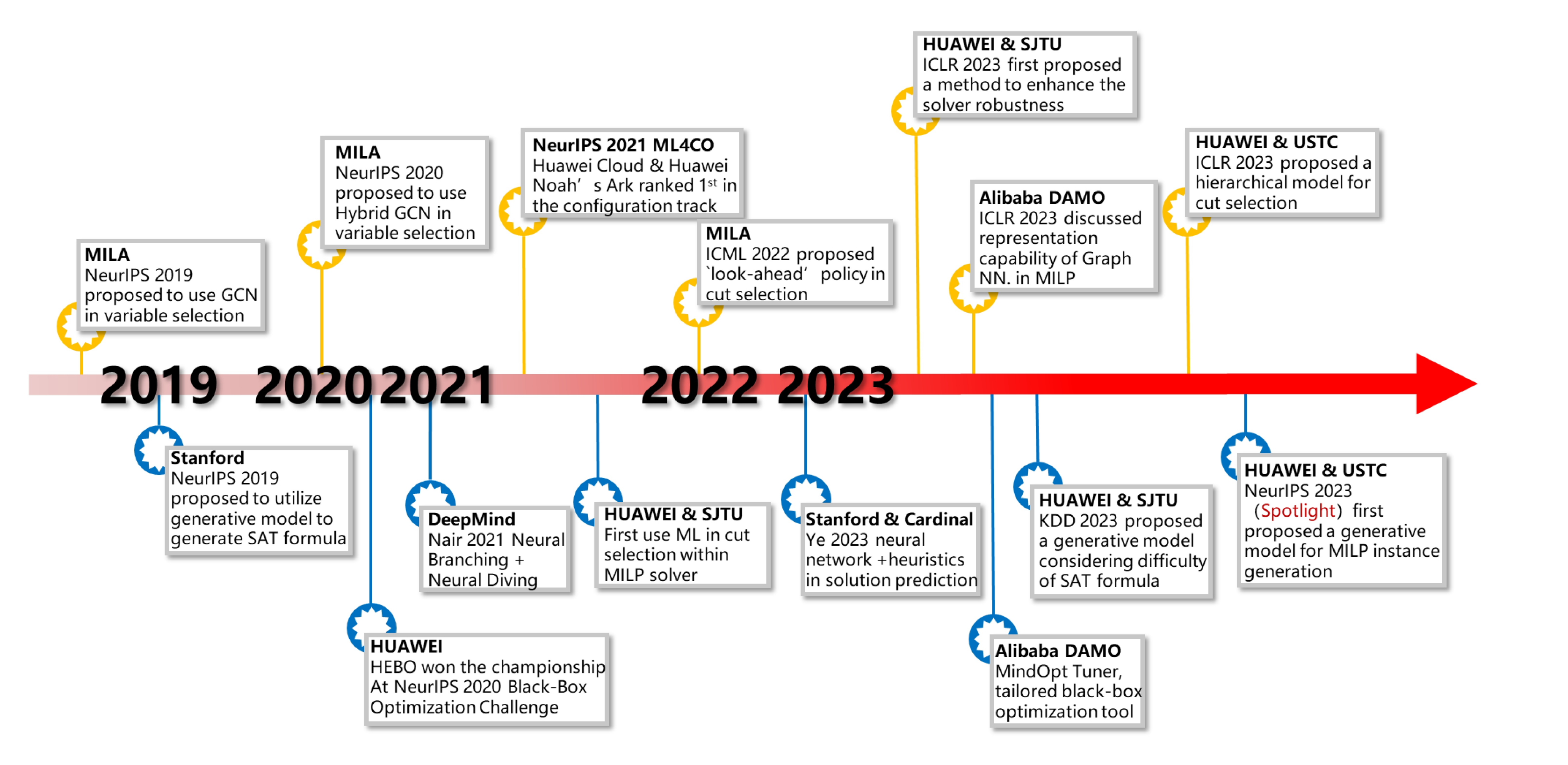}
    \caption{
    The trend of utilizing machine learning techniques to directly solve or to aid in solving the combinatorial problems in recent years. Several seminal works are listed here, especially for data generation, policy learning and hyper-parameter tuning techniques for mathematical programming solvers. Since 2023, this field draws more attentions than ever.
    }
    \label{fig: trend of L2O}
\end{figure}


The integration of machine learning (ML) techniques to address combinatorial optimization (CO) problems marks a transformative trajectory in the intersection of operations research and artificial intelligence. ML's unparalleled ability to discern intricate patterns from data has been leveraged to potentiate solutions for a broad spectrum of CO problems, such as the traveling salesperson problem (TSP) \citep{vinyals2015pointer, bello2016neural, kool2018attention}, satisfiability problem (SAT) \citep{selsam2018learning, yolcu2019learning, guo2023machine}, Directed Acyclic Graph (DAG) scheduling\citep{zhou2022learning}, vehicle routing problem (VRP) \citep{kool2018attention, nazari2018reinforcement, li2021learning,li2018data}, and maximum cut problem (MCP) \citep{barrett2020exploratory}, notably Mixed Integer Linear Programming (MILP) issues \citep{li2023t2tco,li2023accelerating,kuang2023accelerate,bengio2021machine}.

In this manuscript, we mainly focus on fusing ML techniques into mathematical programming solvers. The mathematical programming solver community has witnessed a parallel trend involving the adoption of ML techniques \citep{bengio2021machine, nips21_ml4co_competition,zhang2023survey, guo2023machine}. The seminal works of fusing machine learning techniques into mathematical programming solvers has been depicted in Figure~\ref{fig: trend of L2O}. The efficacy of solvers often hinges on heuristic decisions; thus, the maturation of ML paradigms has begun equipping solvers with enhanced decision-making heuristics \citep{scip_thesis}. Indeed, significant advancements have emerged in critical areas such as selection of cut, variable and node, column generation, and primal heuristics \citep{tang_icml20,l2c_lookahead, adaptive_cut_selection,baltean2019scoring, Khalil_learn_to_branch, pmlr-v80-balcan18a, Parameterizing_branch, learning_to_search, node_uct, morabit2021machine, khalil2017learning,hendel2019adaptive}. In the realm of MIP solver acceleration, GNNs have demonstrated exceptional promise, having been employed in innovative methods for strengthening branching policies and predicting solution variables \citep{ding2020accelerating, gupta2020hybrid, gupta2022lookback, qu2022improved, li2022learning}. The theoretical foundation concerning GNNs' representational capacity for LP also testifies to the method’s expanding applicability \citep{chen2022representing}. These trends attest to the growing confluence of ML with mathematical programming, wherein ML emboldens solvers with advanced heuristic reasoning, ultimately redefining the potency and adaptability of computational problem-solving strategies.



\subsection{Data Generation and Augmentation for Solvers}

\paragraph{Deep Graph Generation} A plethora of literature has investigated deep learning models for graph generation~\citep{guo2022systematic}, including auto-regressive methods~\citep{mercado2021graph}, varational autoencoders (VAEs)~\citep{kipf2016variational}, and generative diffusion models~\citep{fan2023generative}.
These models have been widely used in various fields~\citep{zhu2022survey} such as molecule design~\citep{mahmood2021masked,jin2018junction,gengnovo} and social network generation~\citep{watts1998collective,leskovec2010kronecker}.
G2SAT~\citep{you2019g2sat}, \textit{the first deep learning method for SAT instance generation}, has received much research attention~\citep{li2023hardsatgen,garzon2022performance}.
Nevertheless, it is non-trivial to adopt G2SAT to MILP instance generation, as G2SAT does not consider the high-precision numerical prediction, which is one of the fundamental challenges in MILP instance generation.

\paragraph{MILP Instance Generation} Many previous works have made efforts to generate synthetic MILP instances for developing and testing solvers.
Existing methods fall into two categories.
The first category focuses on using mathematical formulations to generate instances for specific combinatorial optimization problems such as TSP~\citep{vander1995heuristic}, set covering~\citep{Balas1980}, and mixed-integer knapsack~\citep{atamturk2003facets}.
The second category aims to generate general MILP instances.
Bowly~\citep{bowly2019stress} proposed a framework to generate feasible and bounded MILP instances by sampling from an encoding space that controls a few specific statistics, e.g., density, node degrees, and coefficient mean.
However, the aforementioned methods either rely heavily on expert-designed formulations or struggle to capture the rich features of real-world instances.


\paragraph{Adversarial Augmentation on Graphs} Recently, researchers pay more attention to the generalization and robustness of neural solvers \citep{geisler2021generalization, luroco, wang2023asp}.
Geisler et al. \citep{geisler2021generalization} and Lu et al. \citep{luroco} improve the generalization of some problem-specific solvers.
Specifically, Geisler et al. \citep{geisler2021generalization} use adversarial attacks to improve the adversarial robustness of TSP and SAT solvers.
Wang et al. \citep{wang2023asp} combine game theory and curriculum learning to help neural TSP/VRP solvers gradually adapt to unseen distributions and varying scales.
For popular CO solvers, Lu et al. \citep{luroco} propose an adversarial attack approach to evaluate solvers' robustness.
Nonetheless, designing robust practical solvers remains largely unexplored.

\subsection{Policy Learning within Solvers}

The use of graph neural networks (GNNs) to assist optimization solvers has received significant attention in recent years. \citet{gasse2019exact} first proposed a constraint-variable bipartite graph representation for mixed-integer LPs and a GNN model for learning a strong branching policy to accelerate MIP solvers. Following this framework, many researchers have proposed various approaches to improve MIP or LP solvers with GNNs~\cite{ding2020accelerating,gupta2022lookback,qu2022improved,li2022learning}. \citet{gupta2022lookback} discovered a "lookback" property missing in the trained GNN and proposed incorporating it into the training process, which resulted in further speedup for MIP solvers. \citet{qu2022improved} proposed an improved reinforcement learning algorithm that builds upon imitation learning~\cite{gasse2019exact}. Considering that high-end GPUs may not be accessible to many practitioners, \citet{gupta2020hybrid} proposed a hybrid model for branching in MIP inferencing on CPU and maintained competitive speedup. \citet{ding2020accelerating} proposed a tripartite graph representation for MIP and used GNNs to predict solution values for binary variables. \citet{li2022learning} reformulated the LP and reordered the variables and constraints using a GNN and a Pointer Network. More recently, 
\citet{chen2022representing} built a theoretical foundation for the representation power of GNNs for LP, proving that given any LP, a GNN can be constructed that maps from the LP to its feasibility, boundedness, and an optimal solution. Despite the advancements made in using GNNs to assist optimization solvers, there is currently no work on initial basis selection for LP, pushing it towards practicality. 

Besides, cut selection plays an important role in modern MILP solvers \cite{theoretical_cuts, implementing_cutting}. For cut selection, many existing learning-based methods \cite{tang_icml20, l2c_lookahead, cut_ranking} focus on learning which cuts should be preferred by learning a scoring function to measure cut quality. Specifically, \cite{tang_icml20} proposes a reinforcement learning approach to learn to select the best Gomory cut \cite{gomory_cuts}. Furthermore, \cite{l2c_lookahead} proposes to learn to select 
a cut that yield the best dual bound improvement via imitation learning. Instead of selecting the best cut, \cite{cut_ranking} frames cut selection as multiple instance learning to learn a scoring function, and selects a fixed ratio of cuts with high scores. 
In addition, \cite{adaptive_cut_selection} proposes to learn weightings of four expert-designed scoring rules. On the theoretical side, \cite{balcan2021sample, balcan2022structural} has provided some provable guarantees for learning cut selection policies. 

\subsection{Hyperparameter Tuning for Solvers}


Although numerous techniques have been proposed to accelerate the problem solving, the performance of mathematical programming solver still depends on the correct configuration of the solver's hyper-parameters. As the scale of MILP instances to be solved get larger, effective hyper-parameter tuning techniques become crucial to meet the requirement on solving efficiency. However, mathematical programming solvers generally contain hundreds of hyper-parameters controlling the computation of various solving operators. Such a high-dimensional hyper-parameter space and the complex solving process decide that the relationship between a set of hyper-parameter configuration and corresponding solver performance can hardly be modeled explicitly. In other word, the performance function of hyper-parameters is a typical black-box function, the problem of hyper-parameter tuning is actually a black-box optimization (BO) problem.

There are several automatic hyper-parameter tuning softwares were developed to meet the BO demand. \textit{Spearmint}~\citep{spearmint} and \textit{GPyOpt} use Gaussian Processes as the parameter sampling models. Besides, \textit{Hyperopt}~\cite{hyperopt} employs tree-structured Parzen estimator (TPE)~\cite{tpe} and \textit{SMAC} uses random forest which is good at tackling discrete parameter space~\cite{smac}. Recently, Google developed an internal parameter tuning engine termed \textit{Google Vizier}~\cite{vizier}, which defaults to using Batched Gaussian Process Bandits~\cite{gpbandit} to optimize machine learning models and provides several advanced features such as automated early stopping and transfer learning.~\cite{optuna} proposed \textit{Optuna} that allows users to dynamically construct the hyper-parameter space and provides various pruning (i.e., early stopping) algorithms for both parameter searching and  performance estimation. Microsoft developed a distributed parameter tuning framework termed \textit{NNI}, allowing users to improve tuning efficiency via deploying large-scaled distributed parameter estimation~\cite{nni}. Moreover, \textit{NNI} also provided a visualization platform to help users supervising and analyzing the tuning results.

Except for general BO frameworks, some commercial mathematical programming solver also launched their own parameter tuning products, such as the tuner of \textit{Gorubi}~\cite{gurobiTuner}, \textit{Cplex}~\cite{cplexTuner}, and \textit{MindOpt}~\cite{mindoptTuner}. Generally, These solver-oriented tuning tools require one or more instances as the optimizing objective and the parameter estimation is essentially the solving process of the given instances with specific strategy, which is controlled by the recommended hyper-parameters.

\section{Design Principles}

\begin{figure}[t]
    \centering
    \includegraphics[width=1.0\textwidth]{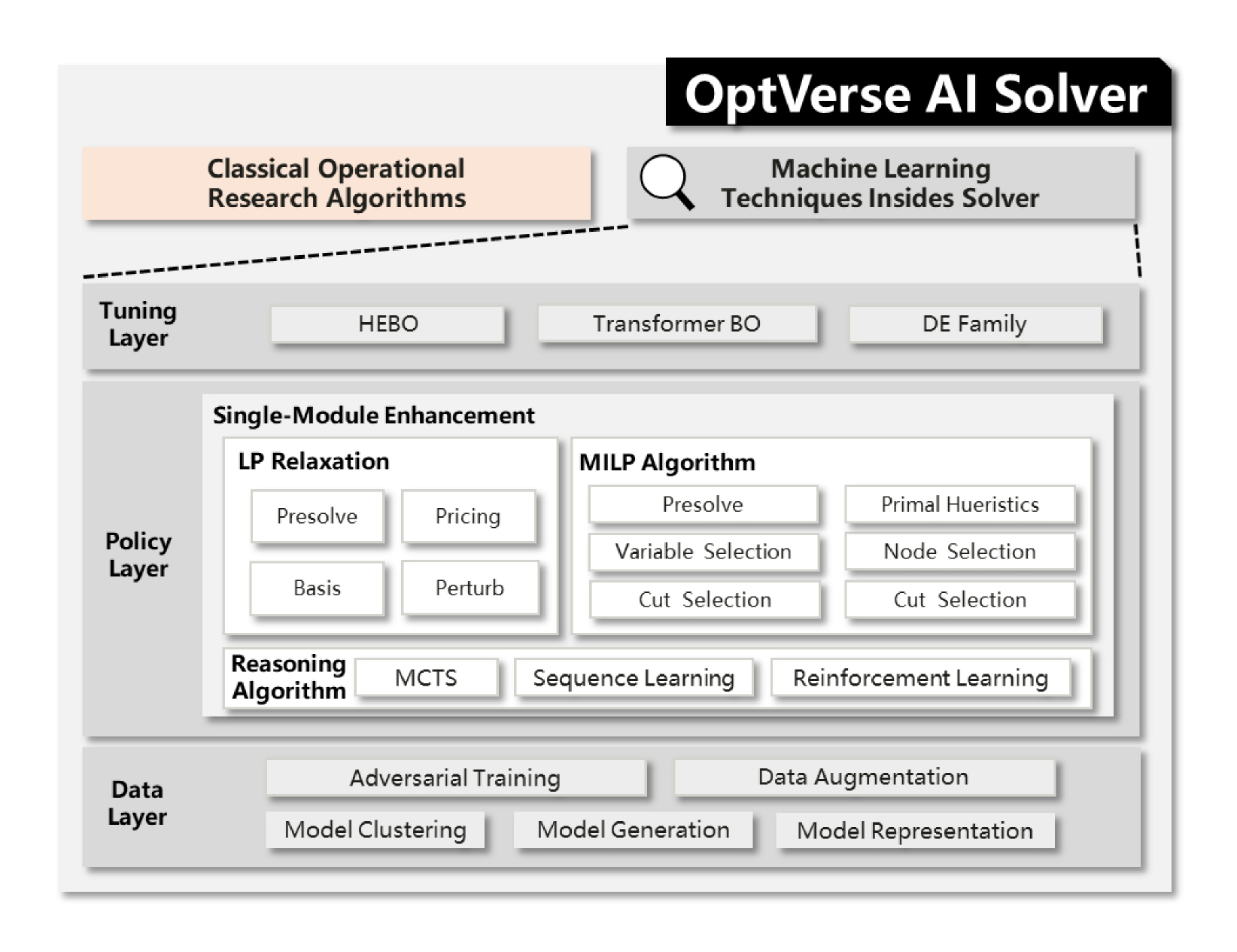}
    \caption{
    The global picture of integrating machine learning techniques into OptVerse AI Solver. There are three main layers of integration, namely, data generation and augmentation for solvers (data layer), policy learning within solvers (policy layer), and parameter tuning for solvers (tuning layer).
    }
    \label{fig:model}
\end{figure}


The integration of machine learning techniques into the OptVerse AI solver is meticulously designed across multiple layers to augment solver efficacy. At the Data Layer, we introduce the ground-breaking HardSATGEN and G2MILP framework for generating Boolean Satisfiability (SAT) and Mixed Integer Linear Programming (MILP) instances respectively, leveraging generative models to replicate real-world problem complexity and synthesize computationally challenging instances. Furthermore, our research mitigates the issue of training learning-based solvers on non-diverse datasets with the introduction of an instance augmentation policy, which employs a contextual bandit problem formulation to adaptively enhance training generalization.

Moving to the Policy Layer, the innovative employment of a Graph Convolutional Network (GCN) streamlines the initial basis selection for the simplex method by capitalizing on the recurring features in LP problems. This is complemented by the first learning-based presolving strategy—RL4Presolve—which orchestrates presolvers using reinforcement learning. We also present a hierarchical sequence model (HEM) for intelligent cut selection in MILP solvers, ensuring an optimized mix of quantity, preference, and sequential ordering of cuts. Moreover, a Neural Diving heuristic is deployed for binary MILP problems, using a GCNN to drive variable assignments that assist in uncovering feasible solutions efficiently.

Finally, in the Tuning Layer, a Parameter Tuning Framework integrates tuners to navigate the hyper-parameter space effectively, underpinning the search with experiments consisting of trials and configurations. This robust design prioritizes both user convenience and search efficiency, with our advanced black-box optimization algorithms, HEBO~\cite{hebo}, Transformber BO~\cite{alex2023end}, and a command-line utility for experiment management and a web server offering rich UI functionality for monitoring experimental metrics and outcomes.

Collectively, these machine learning-oriented strategies—spanning data generation, augmentation, presolving, cut selection, and parameter tuning—formulate a cohesive and intelligent architecture within the OptVerse AI solver, propelling it towards exceptional performance on complex optimization problems.
\section{Applications}


\subsection{Data Generation and Data Augmentation}

The relentless drive to advance the domain of mathematical programming solvers necessitates a wealth of diverse and extensive instance datasets, pivotal for refining solver capabilities through hyperparameter tuning, machine learning (ML) model training, solver evaluation, and invigorating research with challenging benchmarks. Real-world instances are, however, scarce, encumbered by strenuous data curation efforts and proprietary limitations.

To tackle the data scarcity challenge, we propose HardSATGEN~\cite{li2023hardsatgen} and G2MILP~\cite{g2milp1,g2milp2}, aiming to generate more realistic and challenging instances. HardSATGEN specifically tackles the preservation of computational hardness in SAT instances, using a novel one-to-one bipartite graph representation and a split-merge framework enhanced with fine-grained control over community structures and unsatisfiable cores. This approach facilitates more accurate replication of SAT instance hardness. And G2MILP marks the first deep generative framework for MILP instances, leveraging weighted bipartite graphs and Graph Neural Networks (GNNs) to produce instances that are both realistic and computationally challenging. This framework introduces a masked variational autoencoder approach, simplifying the complex graph generation task while preserving essential problem structures~\cite{g2milp1,g2milp2}.

In the context of adversarial training for data augmentation, there is a pressing need for solvers that can generalize to out-of-distribution (OOD) instances. Despite the difficulties of generalization and the restrictions imposed by limited and anonymous data sets, we propose an augmentation policy with a contextual bandit framework to improve solver performance for MILP problems. By approximating the adversarial augmentation task as a contextual bandit problem and utilizing a variant of the Proximal Policy Optimization (PPO) algorithm, we aim to enhance learning efficacy and address the issues of non-differentiability and sparse supervised signals without employing explicit mathematical formulations~\cite{ppo}.

\subsubsection{Data Generation via Generative Models}
\paragraph{Background}
Data, i.e., the collection of MILP and SAT instances plays a fundamental role in developing advanced MILP and SAT solvers, primarily from three aspects.
First, for tasks such as hyperparameter configuration and ML model training, a substantial and diverse set of realistic and independent identically distributed (i.i.d.) instances is necessary.
Second, the evaluation of solvers requires as many instances as possible to identify potential issues and weaknesses through white-box testing.
Finally, to inspire more efficient algorithms and promote research investigations, the research community keeps calling for challenging instances for better benchmarking and competitions.

However, the limited availability of real-world instances, due to labor-intensive data collection and proprietary issues, remains a critical challenge and often leads to sub-optimal decisions and biased assessments.
This challenge motivate a suite of synthetic instance generation techniques, which fall into two categories.
Some methods rely heavily on expert-designed formulations for specific problems, such as Traveling Salesman Problems (TSPs) or Set Covering problems.
However, these methods cannot cover real-world applications where domain-specific expertise or access to the combinatorial structures is limited.
Other methods construct general MILP and SAT instances by sampling from an encoding space that controls a few specific statistics.
However, these methods often struggle to capture the rich features and the underlying combinatorial structures, resulting in an unsatisfactory alignment with real-world instances.
Moreover, these works have scarcely investigated the ability of the generators to produce challenging MILP and SAT instances.

\paragraph{Problem Setup}
We argue that developing deep learning (DL)-based generators is a promising way to address the challenge of limited data availability for the following reasons.
First, these generators can actively learn from real-world instances and generate new ones without relying on expert-designed formulations.
The generated instances can simulate realistic scenarios, cover a wide range of cases, significantly enrich datasets, and facilitate the development of solvers at a relatively low cost.
Second, learning-based generators have the capacity to capture instance features, enabling them to efficiently explore the combinatorial problem space to construct challenging instances.
Finally, this approach shows promising technical prospects for understanding the problem space and learning representations.
Although deep learning techniques have been widely applied to generate Boolean satisfiability (SAT) problems, but the previous work all ignore retaining the difficulty of generated SAT instances. Besides, the development of deep learning-based MILP instance generators remains a complete blank due to higher technical difficulties, i.e., it involves not only the intrinsic combinatorial structure preservation but also high-precision numerical prediction. 

\paragraph{Method: Instances Generation for SAT Solvers}

To enhance the data reliability of SAT instances, we conduct an in-depth analysis of the computational hardness degradation issue in SAT instance generation and propose \textbf{HardSATGEN}, which to our best knowledge, is the first generative model that generates instances maintaining similar computational hardness and capturing the global structural properties simultaneously. We adopt the one-to-one bipartite graph representation for SAT instances, i.e. the literal-clause graph (LCG), which is composed of a literal node set and a clause node set, and an edge therein indicates the occurrence of the literal in the clause. The methodology lies upon the node split-merge framework~\cite{you2019g2sat}, which first gradually splits clause nodes in the bipartite graph to form a forest of template and then enforces a graph neural network (GNN) to learn the reverse node merging process to generate new bipartite graphs. A concrete analysis of the hardness degradation issue for the one-stage node split-merge framework reveals that the root cause lies in the inherent homogeneity in the split-merge procedure and the difficulty of semantic formation of structures led by oversplit substructures. In addition, we observe that industrial formulae exhibit community structure correlated with hardness. Based on the analysis, we introduce a fine-grained control mechanism over the community structure and the unsatisfiable cores (of unsatisfiable instances) to the split-merge paradigm and propose a multi-stage generation pipeline involving scrambling the core, connecting within communities, and connecting cross the communities to guide the iterative split-merge procedure for better similarity in structure and computational hardness. 

\begin{figure*}[!tb]
    \centering
    \includegraphics[width=0.98\linewidth]{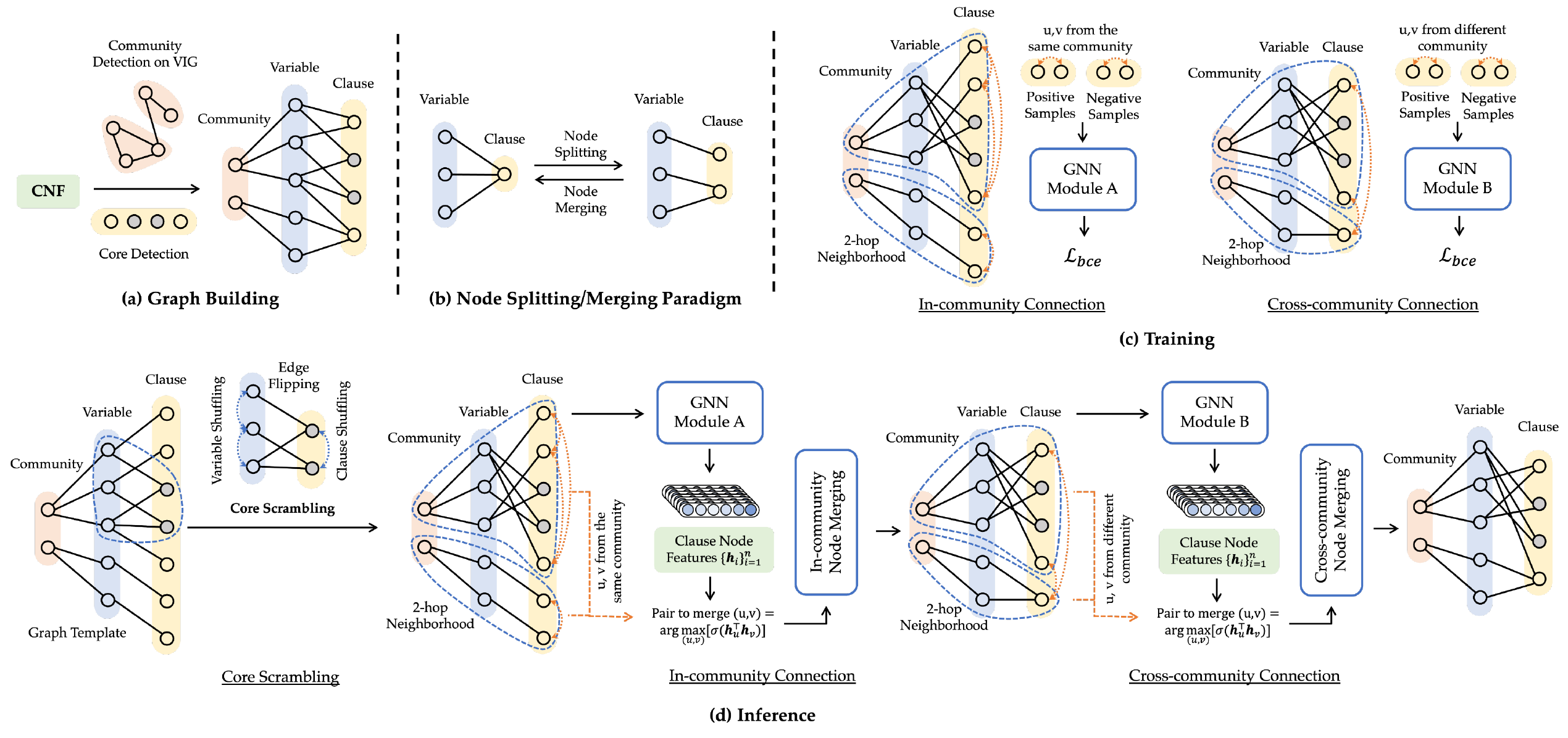}
    \caption{Overview of the HardSATGEN pipeline for SAT instance generation.}
    \label{fig:hardsatgen}
\end{figure*}

Specifically, as shown in Fig.~\ref{fig:hardsatgen}, for an input formula, core detection on the formula and community detection on its variable-incidence graph (VIG) representation, which consists of variables as nodes and the edges indicating the co-occurrence of two variables in one clause, are applied to incorporate knowledge about community and core structures into the graphs. During the splitting phase, we retain the detected core and then remove cross-community connections and the remaining in-community connections to obtain split node pairs, which serve as the training data for two constructive phases, and the remained graph is saved as the templates. In training, two GNNs are trained to correspond to cross-community and in-community connections, learning the structural features and predicting whether the specific node pair should be merged, which is framed as a binary classification task. The trained GNNs are utilized in the inference phase to generate new instances, where the core scrambling operation involves random permutation of variables and clauses, as well as random flipping of literals, and the subsequent in-community and cross-community node merging phases utilize the learned models with set operations on the node sets to construct connections within and cross communities, respectively. For more details about the methodology, please refer to \cite{li2023hardsatgen}.

\paragraph{Method: Instances Generation for MILP Solvers}

\begin{figure*}[!t]
  \centering
  \includegraphics[width=0.48\linewidth]{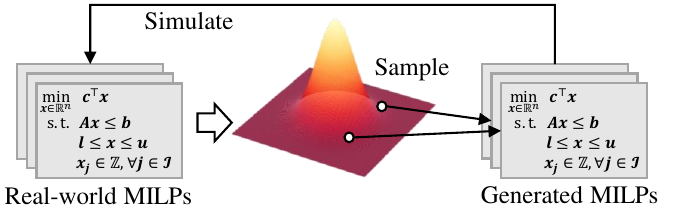}
  \includegraphics[width=0.48\linewidth]{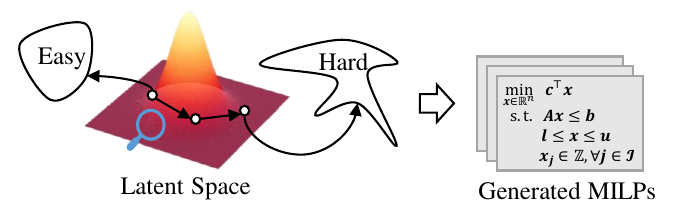}
  \caption{
  We investigate two distinct task settings for MILP instance generation: (left) realistic MILP instance generation and
  (right) hard MILP instance generation.
  }
  \label{fig:g2milp-settings}
\end{figure*}

\begin{figure*}[!tb]
    \centering
    \includegraphics[width=0.9\linewidth]{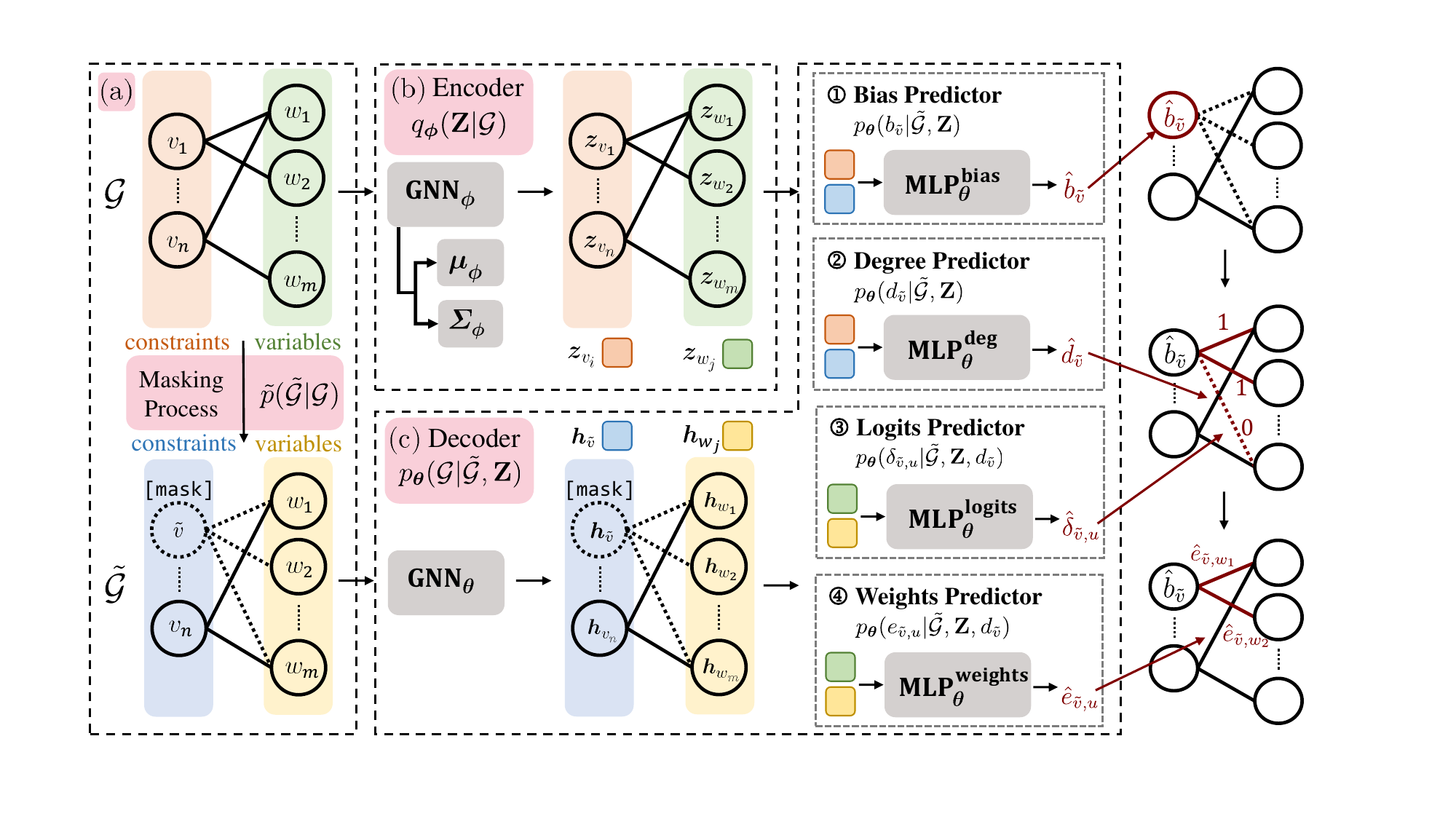}
    \caption{Overview of the G2MILP pipeline for MILP instance generation.}
    \label{fig:g2milp}
\end{figure*}

To enhance the data reliability of MILP instances, we propose {\bf G2MILP}, which to the best of our knowledge, is {\it the first} deep generative framework for MILP instances.
We represent MILP instances as weighted bipartite graphs, where variables and constraints are vertices, and non-zero coefficients are edges.
This graph representation enables us to use graph neural networks (GNNs) to effectively capture the essential features of MILP instances using graph neural networks (GNNs).
In this way, we recast the original task as a graph generation problem.
To accommodate various application scenarios, we consider two task settings for utilizing G2MILP in MILP instance generation: realistic MILP instance generation and hard MILP instance generation, as shown in Figure~\ref{fig:g2milp-settings}.
We begin by focusing on realistic MILP instance generation, where our objective is to generate new MILP instances that closely resemble real-world instances in terms of their structures and computational hardness.
Since generating the complex bipartite graphs from scratch can be computationally expensive and may destroy the intrinsic combinatorial structures of the problems, we propose a masked variational autoencoder (VAE) paradigm inspired by the masked autoencoder (MAE) and the variational autoencoder (VAE) theories.
The model overview is in Figure~\ref{fig:g2milp}.
Specifically, this paradigm iteratively corrupts and replaces parts of the original graphs using sampled latent vectors.
To implement the complicated generation steps, we design a decoder consisting of four modules that work cooperatively to determine multiple components of new instances, encompassing both structure and and numerical prediction tasks simultaneously.
Subsequently, we work on hard MILP instance generation, where our goal is to construct challenging MILP instances.
To achieve this, we propose a hardness-oriented iterative augmenting scheme.
In each iteration, G2MILP generates a batch of new instances and stores the most difficult instances in a storage.
We then fine-tune G2MILP using the instances in the storage as a training set, so that the model is specifically oriented towards generating challenging instances. For more details about this technique, please refer to~\cite{g2milp1,g2milp2}.

\subsubsection{Data Augmentation under Adversarial Training}
\paragraph{Background} Generalizing learning-based solvers to out-of-distribution (OOD) instances poses significant challenges for real-world applications.
First, the solvers are often trained on a given dataset and need to process instances of varying sizes in real-world scenarios.
The sizes of instances encountered in practice can be much larger than those in the training dataset.
Second, the environment changes often introduce perturbations to instance structures.
For example, in staff scheduling problems, emergencies may arise and impose additional restrictions that are not included in the previous MILP problems.
We thus need to consider additional kinds of constraints representing these restrictions, leading to changes in the problem structures.
Third, acquiring or generating more instances from other distributions or with larger sizes may not be feasible due to information security/privacy concerns or high data acquisition costs.
For example, in a company's job-shop scheduling problems, confidential business information, such as the production capacity and costs, can be inferred from the MILP instances.
Thus, the company may provide only a few anonymous training instances to avoid privacy disclosure.

The solving time of the learning-based solvers under the aforementioned settings could be much longer compared to traditional heuristics, particularly when the training instances are few.  
Existing work on the generalization ability of CO solvers mainly focuses on the problem-specific approximation solvers (such as the neural solvers for routing problems) instead of the general exact MILP solvers.
For the problem-specific solvers, we can get access to the problem type and then employ domain-randomization type approaches by generating a large number of instances from different distributions \cite{wang2023asp, geisler2021generalization}.
However, data generation has severe limitations when it comes to anonymous datasets where prior knowledge of the problem type is unavailable.
Consequently, the problem-specific generation algorithms cannot be applied in such cases.
Therefore, we thus focuses on the restrictive practical setting where training data only comes from limited training domains, and acquiring or generating new instances within a short time is intractable.
This challenge significantly hinders the applicability of learning-based solvers to real-world scenarios.

\paragraph{Problem Setup} The instance augmentation policy, denoted as $\Phi:\mathcal{G}\to\mathcal{G}$, maps an instance bipartite graph $G$ to the augmented instance graph $\Phi(G)$ by applying augmentation operators.
However, two challenges for the instance augmentation remain to address.
{First, employing structural augmentation on a bipartite graph leads to a non-differentiable learning objective for the augmentation policy, making it intractable to train through backward propagation and gradient-based updates.}
{Second, the augmentation policy receives significantly less training data (instances graphs) compared to that of the branching policy (branching samples), resulting in low training efficiency for the augmentation policy}.
The adversarial augmentation policy operates on MILP instance graphs $\{G_k\}_k$ to generate augmented instances $\{\Phi(G_k)\}_k$, and each augmented instance can generate a sequence of branching samples. 

Addressing the issues of non-differentiability and unsatisfactory training efficiency requires a sample-efficient training algorithm under sparse supervised signals. 
Thus, we formulate the instances augmentation process as a contextual bandit problem and adopt a contextual-bandit version of the proximal policy optimization (PPO) algorithm to train the augmentation policy \cite{ppo}.

\paragraph{Method} 
To formulate the graph augmentation as a contextual bandit problem, we view the solver as the environment and the adversarial augmentation policy as the agent.
The contextual bandit problem can be represented as the tuple $(\mathcal{G}, \mathcal{A}^{\text{CB}}, r^{\text{CB}})$.
To distinguish the notations from the branching MDP, here we use the superscript CB to specify the contextual bandit formulation.
Then we specify the state space ${\mathcal{G}}$, the action space $\mathcal{A}^{\text{CB}}$ and reward function $r^{\text{CB}}:{\mathcal{G}}\times\mathcal{A}^{\text{CB}}\to\mathbb{R}$ for a MILP instance as follows.
\begin{itemize}
  \item \textbf{The state space} $\mathcal{G}$, or context vector space, also known as the context vector space, is the set of instance graph representations of MILP instances.
  \item
      \textbf{An action} $a^{\text{CB}}=(V^{\text{CB}}, E^{\text{CB}}, W^{\text{CB}})\in\mathcal{A}^{\text{CB}}$, consists of subsets of variables $V^{\text{CB}}$, edges $E^{\text{CB}}$, and constraints $W^{\text{CB}}$ to be masked.
      The agent applies action $a^{\text{CB}}$ to an instance graph $G$ to obtain an augmented instance $\Phi(G)$, on which branching samples are collected.
  \item \textbf{The reward} The reward function $r^{\text{CB}}(G, a^{\text{CB}})$ is defined as the average loss of the branching policy $F(\pi_\theta,\Phi(G),\mathcal{T})$ using the samples collected from the augmented graph $\Phi(G)$ derived from $G$.
\end{itemize}
We describe the training process for the augmentation policy network $\Phi_{\eta}$ with parameters $\eta$ under the contextual bandit formulation as follows.
First, the augmentation policy network $\Phi_{\eta}$ takes as input an instance graph $G$ and outputs the masking probability of each node and edge.
An action $a^{\text{CB}}$ is constructed by selecting the nodes and edges that are most likely to be masked based on the predicted probability and a predefined masking proportion.
We denote the predicted probability of the selected action $a^{\text{CB}}$ by $P_{\eta}(a^{\text{CB}}|G)$.
Second, we leverage a learnable state-value function $V_{\alpha}:\mathcal{G}\to\mathbb{R}$ with parameters $\alpha$ to calculate the variance-reduced estimation of the advantage function $\hat{A}=r^{\text{CB}}-V_{\alpha}(G)$.
Third, to improve the training efficiency of the augmentation policy, we use a replay buffer $\mathcal{B}$ consisting of the tuples $(G,\hat{A},P_{\eta}(a^{\text{CB}}|G)$ obtained from the history augmentation policy for experience replay.
We use $\lambda(\eta)=\frac{P_{\eta_{\text{current}}}(a^{\text{CB}}|G)}{P_{\eta_{\text{old}}}(a^{\text{CB}}|G)}$ to represent the likelihood ratio between the current policy and old policies sampled from the buffer.
Finally, the training objective for the adversarial augmentation policy, which includes the policy network and state-value function, is achieved by minimizing the following equations:
\begin{align}\label{PPOobj}
L_{V}(\alpha)&=\mathbb{E}_\mathcal{B}\left[\hat{A}^2\right]\quad\\
L_{\Phi}(\eta)&=\mathbb{E}_\mathcal{B}\left[\min(\lambda(\eta), \text{clip}(\lambda(\eta),1-\epsilon, 1+\epsilon))\hat{A})\right],
\end{align}
where $\epsilon$ is the clip ratio.

For more details about this technique, please refer to~\cite{liu2023promoting}.

\paragraph{Deployment and Performance Gain} To evaluate the performance in real-world applications, we deploy AdaSolver to a real-world Anonymous dataset, a widely-used subset of the MILPLIB library.
The number of constraints in the Anonymous dataset ranges from 3,375 to 159,823, and the number of variables ranges from 1,613 to 92,261.
The wide range of problem sizes as well as real-world perturbations pose significant challenges for the learning-based methods.
The results in Table \ref{table:real_world_dataset} demonstrate that AdaSolver is able to improve the performance of both IL- and RL-based solvers.
AdaSolver-IL achieves the best PD gap while AdaSolver-RL achieves the best PD integral metric.
The performance of AdaSolver sheds light on the generalizable learning-based solvers in real-world applications.

\begin{table*}[t]
\caption{The results in the challenging real-world Anonymous dataset show that AdaSolver-IL achieves the lowest PD gap and AdaSolver-RL achieves the lowest PD integral.}
\label{table:real_world_dataset}
\centering
\scalebox{.9}{
\begin{tabular}{@{}ccccc@{}}
\toprule
Method & Time(s) & Nodes  & PD integral $\downarrow$ & PD gap $\downarrow$\\\midrule
RPB & 866.68 & 43907.95& 76018.92& 5.50e+19\\\midrule
SVMRank &899.93&14519.66&65522.04& 4.83e+19 \\
Trees &900.00& 9658.55&76143.65&4.66e+19\\
LMART &830 &16229.83&62415.32&4.66e+19\\
tMDP+DFS &763.81 & 40933.30&44679.11&5.00e+18 \\
GNN & 823.88& 37469.58&45483.01& 3.50e+18 \\\midrule
AdaSolver-RL&754.08&46746.05 &\textbf{42179.25}&5.00e+18\\
AdaSolver-IL & 830.50 & 57156.53 & 42579.31 & \textbf{3.33e+18}\\ \bottomrule
\end{tabular}
}

\end{table*}

\subsection{Policy Learning within Solver}


In the evolving sphere of mathematical programming solvers, policy learning emerges as an indispensable tool, primarily due to its penchant for capturing sophisticated patterns and adaptively enhancing decision-making processes that are considerably nuanced for conventional rule-based systems. The imperative for such AI-driven policies originates from the need to confront the intricacies and diversities inherent in linear and mixed integer programming problems, where efficiency gains can translate to significant computational savings. AI policies augment solvers with the cognitive dexterity to personalize strategies for individual problem instances, thereby boosting convergence rates and ameliorating overall solver performance.

For Initial Basis Selection in solving LP problems, the employment of a Graph Convolutional Network (GCN) stands as a testament to the viability of machine learning in devising efficacious initial bases by exploiting recurrent traits in problem instances. In the quest to eradicate redundancies within LP constraints during the Presolving phase, the machine learning-centric innovation of reinforcement learning for presolve (RL4Presolve) tactically sequences presolvers to significantly compress the cumulative decision timeline. When navigating the prolific domain of Cut Selection in MILP, a hierarchically-structured sequence model (HEM) meticulously curates a subset of cuts that synergistically bolsters the dual bound efficacy, once again reflecting a sagacious coupling of policy learning with operational solver optimization. Diving deeper into MILP, the Neural Diving heuristic exemplifies the ingenuity of employing neural networks, specifically a GCNN, to precipitate the uncovering of feasible solutions via the discrete curation of binary variables assignments, showcasing a parallelizable and scalable machine learning application.

Each of these strategies ingrains a sophisticated artifice within the solvers, channeling the potent blend of machine learning and operational research to elevate computational performance. The discourse on machine learning policies points to a burgeoning era where mathematical programming transcends its conventional boundaries, realizing enhanced precision and efficiency through the keen insights rendered by artificial intelligence.

\subsubsection{Initial Basis Selection}

\begin{figure}[t]
    \centering
    \includegraphics[width=.9\textwidth]{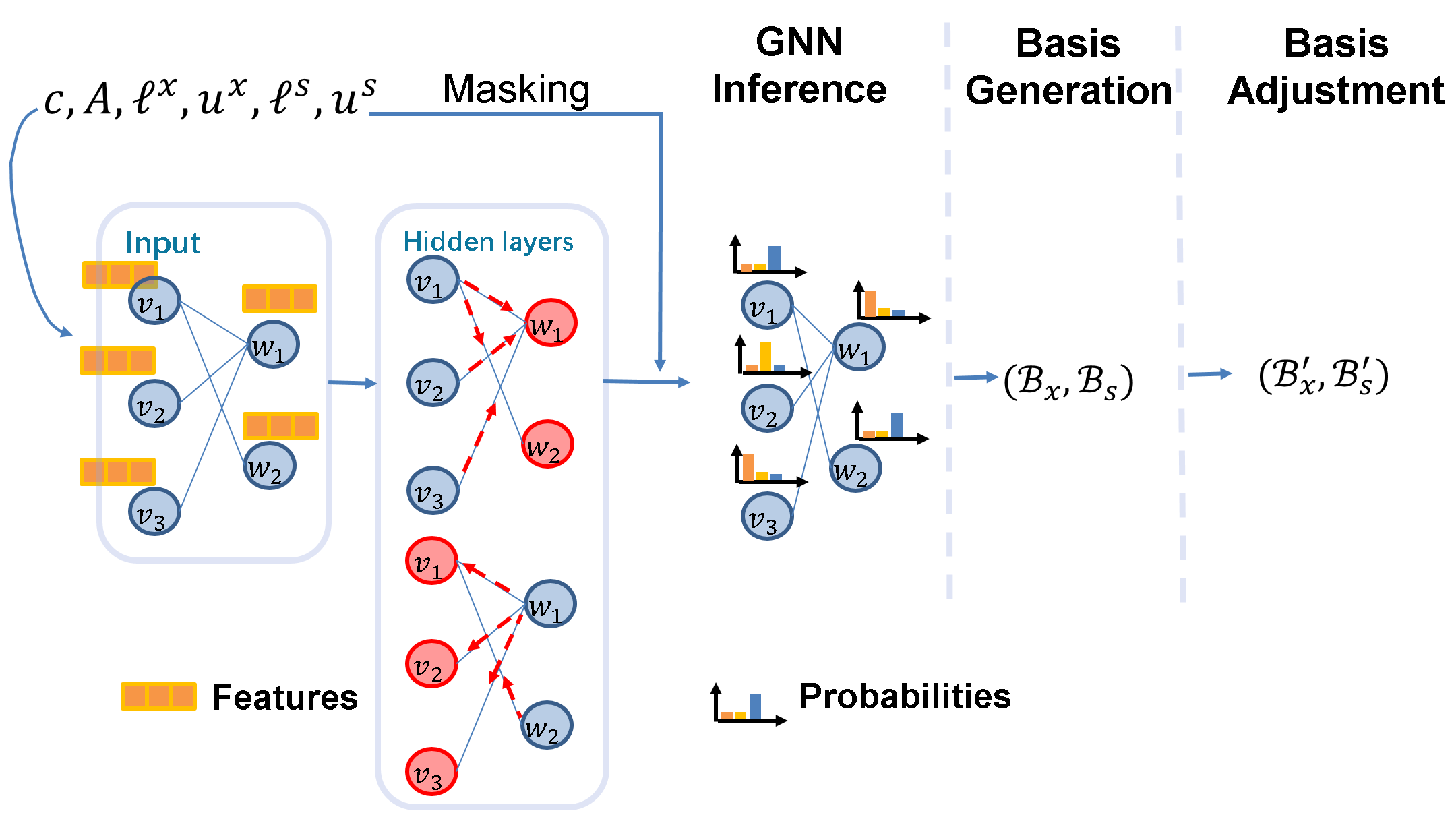}
    \caption{Overall procedure of the inference steps in our proposed initial basis selection method. 
    }
    \label{fig:gnn-arch}
\end{figure}

\paragraph{Background} The Simplex method \cite{dantzig1960decomposition} is a pioneering method for solving large-scale LP problems. It starts with {initial basis} $\mathcal{B}^{(0)}$ and routinely pivots to a neighboring basis with improvement till reaching an optimal basis $\mathcal{B}^{*}$. 
The performance of the Simplex method largely depends on the choice of initial basis. 
Current strategies 
\citet{bixby1992implementing}, \citet{gould1989new}, \citet{junior2005improved}, \citet{ploskas2021triangulation}, \citet{galabova2020idiot}
for selecting an advanced initial basis are mostly rule-based, requiring extensive expert knowledge and empirical study. 
Yet, most of them still fail to exhibit consistent improvement over the canonical initial basis, because they often rely on heuristics that do not generalize well.
Nevertheless, the quest for universally effective strategies may not be necessary in practice, as we often encounter sets of LP problems that exhibit considerable similarities. For instance, in scenarios like a manufacturer's daily production planning or an airport's hourly flight scheduling, similar LP problems regularly occur. 
Therefore, it is natural to ask if a learning-based approach could be more effective, leveraging the correlation among LP problems and enhancing the efficiency of the Simplex method. 

\paragraph{Problem Setup} Denote the set of past solved Linear Programs (LPs) as $\mathcal{D} = \left\{ [(P^k), (x^k, s^k)] \right\}_{k=1}^K$, where $(P^k)$ is the $k$-th problem and $(x^k,s^k)$ is its optimal solution. 
Given $\mathcal{D}$ and a new LP $P^{test}$ from the same distribution, the goal is to predict an initial basis for the new LP. This basis should enable the Simplex algorithm to converge more rapidly, taking advantage of the inherent similarities within the set \(\mathcal{D}\). 
We consider the following standard format of LP: 

\begin{equation}
\begin{array}{cl}
\min _{x \in \mathbb{R}^n, s \in \mathbb{R}^m} & c^T x \\
\text{s.t.} & Ax = s \\
& \ell^x \leq x \leq u^x \\
& \ell^s \leq s \leq u^s,
\end{array} \tag{P}
\end{equation}
where \( \ell^x, \ell^s \) can reach \( -\infty \), and \( u^x, u^s \) can reach \( \infty \). 

This problem presents several challenges: (1) LP problem $P^k$ is of varying size. (2) For each LP problem, the number of potential bases is exponential w.r.t the problem size. (3) The selected basis has to be valid, i.e., the corresponding basis matrix is non-singular and the states of non-basic variables are consistent with their bounds. 

\paragraph{Method} We represent a linear programming (LP) problem as a bipartite graph, as suggested by  \cite{gasse2019exact}. This approach allows for the effective handling of LP problems of varying sizes and complexities. Furthermore, we employ a 2-layer Graph Convolutional Network (GCN), following the model proposed by \cite{morris2019weisfeiler}, to predict the optimal basis for an LP problem.
The bipartite graph representation effectively captures the relationship between variables and constraints in an LP problem, transforming it into a structure that can be efficiently processed by the GCN. The 2-layer GCN then processes this graph to predict the status of each variable in the LP problem, classifying them into categories such as basic, non-basic at the upper bound, or non-basic at the lower bound. This prediction is essential for selecting an appropriate initial basis for the simplex method. In this way, we transform the complex task of initial basis selection into a more manageable classification task. 

To ensure the predicted status of each variable is consistent with its bounds in the LP formulation, a knowledge-based masking technique \cite{fan2022knowledge} is employed. This technique masks out probability entries based on the problem's knowledge, ensuring that the resulting probabilities satisfy the feasibility of non-basic entries. 
During the inference phase, after generating a candidate basis from the predicted probabilities, the basis needs to be adjusted to ensure it is valid, i.e., the matrix \([A_{\mathcal{B}_x}\enspace -I^m_{\mathcal{B}_s}]\) is non-singular. This adjustment is inspired by a basis repair procedure, which involves factorization steps and, if necessary, removing columns that do not meet certain criteria to achieve a complete and successfully factored basis. Figure \ref{fig:gnn-arch} shows the inference steps of the proposed method. Please refer to \cite{pmlr-v202-fan23d} for more details. 




\begin{table*}
\caption{Performance comparison between the rule-based initial-basis and proposed GNN strategy, with the dual simplex method and the OptVerse AI solver. 
} 
\centering 
\scalebox{.9}{

\begin{tabular}{c l c c c c c}
    \toprule
     & {} & \multicolumn{5}{c}{Time ($s$)} \\ \cmidrule{3-7}
   Index & {Dataset} & DEFAULT & CA & CA-MPC & CA-ANG & GNN(Ours) \\ 
    \midrule
   1&  LIBSVM & $16.6{\scriptscriptstyle \pm 10.0}$ & $16.7{\scriptscriptstyle \pm 10.0}$ & $27.9{\scriptscriptstyle \pm 12.4}$ & $28.3{\scriptscriptstyle \pm 2.2}$ & $\bf 11.0{\scriptscriptstyle \pm 3.7}$\\ 
   2& MIRP & $22.1{\scriptscriptstyle \pm 23.3}$ & $21.4{\scriptscriptstyle \pm 22.5}$ & $18.6{\scriptscriptstyle \pm 16.9}$ & $21.6{\scriptscriptstyle \pm 20.9}$ & $\bf 15.4{\scriptscriptstyle \pm 15.7}$\\
  3&  StochasticSupplyChain & $44.6{\scriptscriptstyle \pm 11.8}$ & $61.3{\scriptscriptstyle \pm 12.3}$ & $51.3{\scriptscriptstyle \pm 12.4}$ & $53.2{\scriptscriptstyle \pm 8.5}$ & $\bf 42.7{\scriptscriptstyle \pm 30.0}$\\ 
   4& Generated & $1.3{\scriptscriptstyle \pm 0.2}$ & $1.4{\scriptscriptstyle \pm 0.2}$ & $1.4{\scriptscriptstyle \pm 0.3}$ & $1.4{\scriptscriptstyle \pm 0.3}$ & $\bf 0.5{\scriptscriptstyle \pm 0.5}$\\   
    5&SupplyChain (Small) & $77.9{\scriptscriptstyle \pm 68.4}$ & $85.8{\scriptscriptstyle \pm 80.3}$ & $86.1{\scriptscriptstyle \pm 79.5}$ & $100.1{\scriptscriptstyle \pm 94.0}$ & $\bf 22.8{\scriptscriptstyle \pm 23.5}$\\
    6&SupplyChain (Medium) & $348.7{\scriptscriptstyle \pm 101.0}$ & $1.3K{\scriptscriptstyle \pm 698.2}$ & $382.8{\scriptscriptstyle \pm 102.3}$ & $338.7{\scriptscriptstyle \pm 181.5}$ & $\bf 87.3{\scriptscriptstyle \pm 25.4}$\\
    \bottomrule 
\end{tabular}

}
\label{tab:eval-optv-dual}
\end{table*}

\paragraph{Deployment and Performance Gain} It is crucial that the LP problems solved using this method are similar in nature. This similarity ensures that the learned model can effectively generalize and apply its learned knowledge to new but related LP problems. Our method works best when applied to LP instances that share characteristics with those in the training set.
In practice, the deployment steps could be: 
\begin{enumerate}
    \item \textbf{Initial Phase:}
    In the initial phase, users upload LP problems which are solved completely using the default strategy to accumulate past solving experience.
    As LP problems are solved, their data (including the problems themselves and their solutions) are stored. 
    This phase is critical for accumulating a diverse and representative training set, which is essential for training an effective GNN model.
 
    \item \textbf{Training the GNN Model:}
    When enough correlated data is accumulated, there is an option to start training the GNN model. Actually, it remains difficult to decide when to start training. In practice, we can divide the data into training and validation sets. The performance of trained GNN  is closely monitored on the validation set. If the validation performance indicates a significant acceleration of the simplex algorithm, it implies the time to start training.

    \item \textbf{Applying the Method to Future LPs:}
    For future LP problems provided by the user, the trained GNN model is employed. The model accelerates the solving process by predicting an initial basis that is closer to the optimal solution, thereby reducing the number of iterations and overall computation time required for solving the LP.
\end{enumerate}

Through these steps, the proposed method can be deployed to practical scenarios. Its effectiveness is evidenced by the speed enhancements observed in various benchmarks. Table \ref{tab:eval-optv-dual} highlights the end-to-end time reduction achieved using our GNN model across different datasets: MIRP~\cite{papageorgiou2014MIRPlib}, LIBSVM~\cite{zhu20031,applegate2021practical}, StochasticSupplyChain~\cite{castro2020new}, a synthetic dataset (Generated)~\citet{bowly2020generation}, and two Huawei supply chain datasets -- SupplyChain (Small) and SupplyChain (Medium). The effectiveness of the GNN model is contrasted with baseline methods CA~\cite{bixby1992implementing}, CA-MPC~\cite{ploskas2021triangulation}, and CA-ANG~\cite{junior2005improved}, demonstrating its superior performance in accelerating LP problem solving by harnessing past-solving experiences and the GNN's predictive capabilities.

\subsubsection{Presolving}

\begin{figure*}[t]
\centering
\includegraphics[width=1.0\textwidth]{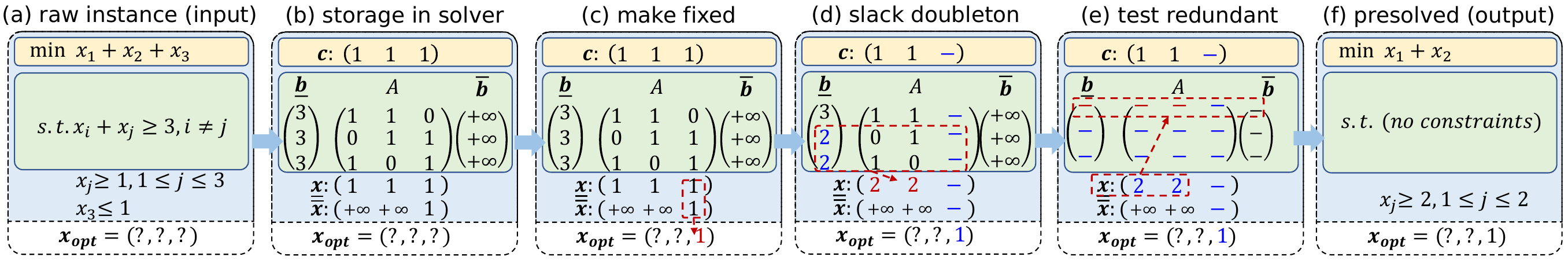}
\caption{A simple example of the presolve process in the modern LP solver. The solver executes different presolvers iteratively to remove the redundancy. 
The red elements are modified by the current presolver and the blue ones by previous presolvers. 
Note that the activation conditions of the current presolver (red box) usually rely on the effect of previous presolvers (blue elements).}
\label{fig:rl4presolve/example}
\end{figure*}

\paragraph{Background} Presolve simplifies input LP problems by equivalent transformations. It is widely recognized as one of the most crucial components in modern linear programming (LP) solvers \cite{ct,ce}. 
In practice, we found LP problems from industry usually contain much redundancy, which could come from reasons like general purpose modeling systems and non-expert modelings, and it can severely decrease the efficiency and reliability of LP solvers to solve LPs \cite{ct}. 
There are many types of redundancy, including multiple constraints (rows) that are linear dependent with each other, a single variable (column) whose value can be fixed in advance \cite{apt}, etc.
Modern LP solvers integrate a rich set of presolvers to handle the redundancy from different aspects \cite{plp}. 
Within OptVerse AI solver, we employ eighteen different presolvers to handle various redundancy from different aspects. 
An illustrate of the presolve process with a toy example can be found in Figure \ref{fig:rl4presolve/example}. 

\paragraph{Problem Setup} In practice, we found that presolve routines, i.e., a program determining a sequence of presolvers successively executed in the presolving phase, play \textit{critical} roles in the efficiency of solving LPs. Previous research on designing high performance presolve routines is relatively limited. Thus, we empirically conclude three points in presolve routine design. That is, which presolvers to select, in what order to execute, and when to stop. Designing such presolve routines is challenging due to the enormous search space and the extensive requirement for expert knowledge. Moreover, hard-coded routines integrated in many modern solvers \cite{clp, highs} cannot capture distinct patterns of different problems to achieve high performance.

\begin{figure}
\centering
\includegraphics[width=0.62\textwidth]{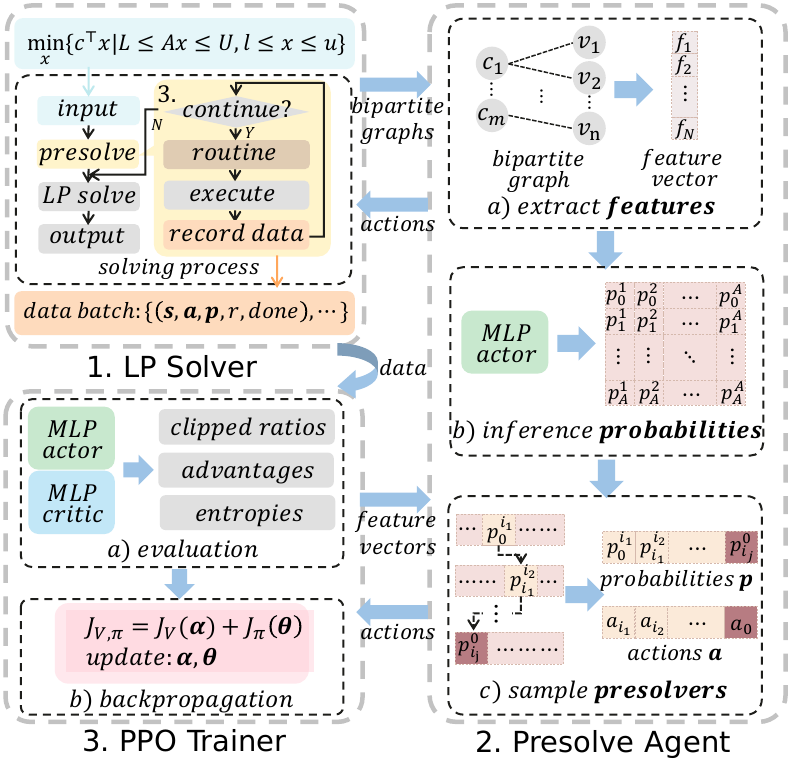}
\caption{Illustrate all modules in RL4Presolve. Here $N,A$ are the length of features and the total number of presolvers, $p_i^j = \pi_{\boldsymbol{\theta}}(i\mid j, \mathbf{s})$ is the conditional probability from $i$th presolver to $j$th, and $a_0=E$ is the end token.}
\label{fig:rl4presolve/illustration}
\end{figure}

\paragraph{Method} In the OptVerse AI solver, we propose the first learning-based approach---that is, reinforcement learning for presolve (RL4Presolve)---to accelerate presolve in large-scale LPs. 
Specifically, there are three core components in RL4Presolve (see Figure \ref{fig:rl4presolve/illustration}):
\begin{enumerate}
    \item The Markov decision process (MDP) formulation. We formulate the task as a Markov decision process by defining
    : a) the presolve state as a 51-dimensional feature vector  based on the activation condition of each presolver. b) the action space as the set of all possible presolver sequences. That is, $\mathcal{A} = \left\{ \boldsymbol{a} \mid \boldsymbol{a} = (a_i)_{i=1}^n,\,a_i\in \mathcal{A}_0, n\in \mathbb{N} \right\}$, where $\mathcal{A}_0$ is the set of all available presolvers in OptVerse. c) the environment transition as the problem changes after the presolver sequence is executed. d) the reward function as the executing time between two steps (i.e., $r=-t$). e) the discount factor $\gamma=1$, and thus the absolute value of the cumulative rewards equals to the total solving time of the input problem.
    \item The agent with adaptive action sequence. Compared to traditional reinforcement learning (RL) algorithms, we need to explicitly consider the cumulative decision time. Thus, select a single presolver for each decision can be relatively  expensive. To tackle this problem, we propose a novel approach that replaces primitive presolvers with automatically generated presolver sequences at each step. Specifically, we employ an agent that learns a one-step conditional probability of executing presolver $j$ after presolver $i$ is executed at last step, which is motivated from combos in video games. Then, we can sample a sequence of presolvers from this learned probability matrix at each step. Intuitively, this  is an automatic way to generate high-quality ``combos'' as each step.
    \item Training algorithm with proximal policy optimization (PPO). Since evaluating the learned presolve routines is time-consuming, we employ the training efficient on-policy RL algorithm PPO to optimize the one-step conditional probabilities. Specifically, the probability ratio can be written as:
\begin{equation}\label{eq: unclipped-ratio}
r(\boldsymbol{\theta})= \frac{\pi_{\boldsymbol{\theta}}\left(a_1 \mid \mathbf{s}\right)\prod_{i=2}^{n+1} \pi_{\boldsymbol{\theta}}\left(a_i \mid a_{i-1}, \mathbf{s}\right) }{\pi_{\boldsymbol{\theta}_{\text{old}}}\left(a_1 \mid \mathbf{s}\right)\prod_{i=2}^{n+1} \pi_{\boldsymbol{\theta}_{\text{old}}}\left(a_i \mid a_{i-1}, \mathbf{s}\right) } = \prod_{i=1}^{n+1} \frac{\pi_{i}(\boldsymbol{\theta})}{\pi_{i}(\boldsymbol{\theta}_{\text{old}})},
\end{equation}
where $\pi_{1}(\boldsymbol{\theta})=\pi_{\boldsymbol{\theta}}(a_1\mid \boldsymbol{s})$, $\pi_{i}(\boldsymbol{\theta})=\pi_{\boldsymbol{\theta}}(a_i\mid a_{i-1}, \boldsymbol{s})$ for $ i\geq 2$, and $a_i$ is the $i$th presolver in OptVerse.    
\end{enumerate}

Please refer to \cite{rl4presolve} for more details. 

\begin{table*}[t]
\caption{
Evaluate RL4Presolve on four real-world benchmarks. 
Results show that RL4Presolve and the rules extracted from it significantly and consistently improves the efficiency of solving LPs.
} 
\centering
\small
\begin{tabular}{ccccc}\toprule
Dataset:                             & \multicolumn{2}{c}{Master Production Schedule} & \multicolumn{2}{c}{Production Planning} \\ \midrule
Method                               & Time(s)             & Improvement(\%)            & Time(s)        & Improvement(\%)        \\ \midrule
Default                              & 3.70                & NA                         & 5.99           & NA                     \\
Expert Tuning                        & 3.10                & 16.20                      & 3.83           & 35.98                  \\
RL4Presolve                          & 2.02                & 45.41                      & 2.81           & 53.04                  \\
Extracted Rule & 2.29                & 38.20                      & 3.25           & 45.73 \\ \bottomrule                
\end{tabular}

\begin{tabular}{ccclc} \toprule
Dataset:                             & \multicolumn{2}{c}{Supply Demand   Matching}          & \multicolumn{2}{c}{MIRPLIB}                             \\ \midrule
Method                               & Time(s)                   & Improvement(\%)           & \multicolumn{1}{c}{Time(s)} & Improvement(\%)           \\ \midrule
Default                              & 17.89                     & NA                        & 115.79                      & NA                        \\
Expert Tuning                        & 13.32 & 25.54 & 104.14 & 10.06 \\
RL4Presolve & 10.12 & 43.44 & 96.98 & 16.25 \\
Extracted Rule & 10.62 & 40.62 & 101.51  & 12.33 \\ \bottomrule                   
\end{tabular}
\label{tab:rl4presolve/routine-rule}
\end{table*}

\paragraph{Deployment and Performance Gain} Applying the RL agent directly to real-world applications is usually challenging due to the hardware constraints for high-end GPUs. Thus, we  extract rules from learned policies for simple and efficient deployment to the solver. Specifically, we observe that RL4Presolve tends to generate highly \textit{similar} action probabilities for instances from similar distributions, which might because that independent and identically distributed instances from real-world tasks are usually generated from similar models with different daily data instantiation. Thus, we can extract some new routines from learned policies that outperforms the default one. Specifically, we sample $20$ presolve routines on each task from learned policies and then replace the default routine in OptVerse with the best-performed one. 
We find this approach \textit{consistently} improves the hard-coded presolve routines. We report the improved efficiency of solving large-scale LPs on four real-world benchmarks (three of which from Huawei's supply chain) in Table \ref{tab:rl4presolve/routine-rule}.




\subsubsection{Cut Selection}
\paragraph{Background}
Given a standard form of Mixed Integer Linear Programming (MILP), we drop all its integer constraints to obtain its \textit{linear programming (LP) relaxation}, which takes the form of  
\begin{align}\label{lp_relaxation}
        z_{\text{LP}}^* \triangleq \min_{\textbf{x}} \{\textbf{c}^{\top} \textbf{x} | \textbf{A}\textbf{x} \leq \textbf{b},\textbf{x}\in \mathbb{R}^n\}.
\end{align}
    Since the problem in (\ref{lp_relaxation}) expands the feasible set, we have $z_{\text{LP}}^* \leq z^*$, where $z^*$ is the optimal solution to the given MILP. We denote any lower bound found via an LP relaxation by a \textit{dual bound}.
    Given the LP relaxation in (\ref{lp_relaxation}), cutting planes (cuts) are linear inequalities that are added to the LP relaxation in the attempt to tighten it without removing any integer feasible solutions of given MILP. Cuts generated by MILP solvers are added in successive rounds. Specifically, each round $k$ involves (1) solving the current LP relaxation, (2) generating a pool of candidate cuts $\mathcal{C}^k$, (3) selecting a subset $\mathcal{S}^k\subseteq \mathcal{C}^k$, (4) adding $\mathcal{S}^k$ to the current LP relaxation to obtain the next LP relaxation, (5) and proceeding to the next round.
    Adding all the generated cuts to the LP relaxation would maximally strengthen the LP relaxation and improve the lower bound at each round. However, adding too many cuts could lead to large models, which can increase the computational burden and present numerical instabilities \citep{implementing_cutting}. Therefore, 
    cut selection is proposed to select a proper subset of the candidate cuts, which is significant for improving the efficiency of solving MILPs \citep{tang_icml20}.

\paragraph{Problem Setup}
Cut selection heavily depends on \textbf{(P1)} which cuts should be preferred, and \textbf{(P2)} how many cuts should be selected \citep{scip_thesis, theoretical_cuts}. Moreover, we observe from extensive empirical results that \textbf{(P3)} what \textit{order of selected cuts} should be preferred significantly impacts the efficiency of solving MILPs as well. To improve the efficiency of MILP solvers, recent methods \citep{tang_icml20, l2c_lookahead, cut_ranking} propose to learn cut selection policies via machine learning, especially reinforcement learning. They offer promising approaches to learn more effective heuristics by capturing underlying patterns among MILPs from specific applications. However, many existing learning-based methods \citep{tang_icml20, l2c_lookahead, cut_ranking}---which learn a scoring function to measure cut quality and select a fixed ratio/number of cuts with high scores---suffer from two limitations. First, they learn which cuts should be preferred by learning a scoring function, neglecting the importance of learning the number and order of selected cuts \citep{theoretical_cuts}. Second, they do not take into account the interaction among cuts when learning which cuts should be preferred, as they score each cut \textit{independently}. As a result, they struggle to select cuts that complement each other nicely, which could severely hinder the efficiency of solving MILPs \citep{theoretical_cuts}. Indeed, we empirically show that they tend to select many similar cuts with high scores. 

\begin{figure}[t]
\centering
\includegraphics[width=0.65\textwidth]{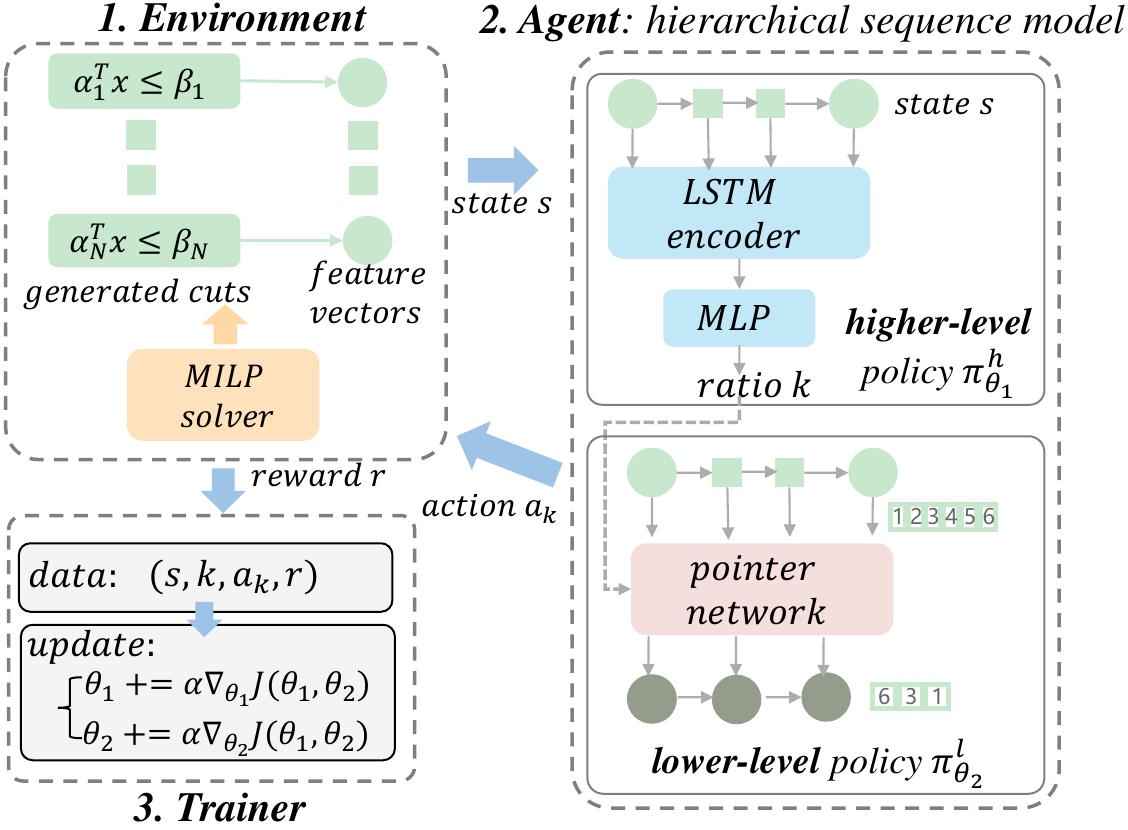}
\caption{Illustrate all modules in HEM. We formulate a MILP solver as the environment and the HEM as the agent. Moreover, we train HEM via a hierarchical policy gradient algorithm.}
\label{fig:hem/illustration}
\end{figure}

\paragraph{Method}
In the OptVerse AI solver, we propose a novel \textbf{h}ierarchical s\textbf{e}quence \textbf{m}odel (HEM) to learn cut selection policies via reinforcement learning \citep{hem}, which is \textit{the first} learning-based method that can tackle \textbf{(P1)-(P3)} simultaneously. Specifically, there are three core components in HEM (see Figure \ref{fig:hem/illustration}):
\begin{enumerate}
    \item \textbf{Reinforcement Learning Formulation} As shown in Figure \ref{fig:hem/illustration}, we formulate a MILP solver as the environment and our proposed HEM as the agent. 
    We consider an MDP defined by the tuple
    $(\mathcal{S}, \mathcal{A}, r, f)$.
    Specifically, we specify the state space $\mathcal{S}$, the action space $\mathcal{A}$, the reward function $r:\mathcal{S} \times \mathcal{A} \to \mathbb{R}$, the transition function $f$, and the terminal state in the following. 
    \textbf{(1) The state space $\mathcal{S}$.}  
    Since the current LP relaxation and the generated cuts contain the core information for cut selection, we define a state $s$ by $(M_{\text{LP}}, \mathcal{C}, \textbf{x}_{\text{LP}}^*)$. Here $M_{\text{LP}}$ denotes the mathematical model of the current LP relaxation, $\mathcal{C}$ denotes the set of the candidate cuts, and $\textbf{x}_{\text{LP}}^*$ denotes the optimal solution of the LP relaxation. To encode the state information, we design thirteen features for each candidate cut based on the information of $(M_{\text{LP}}, \mathcal{C}, \textbf{x}_{\text{LP}}^*)$. That is, we actually represent a state $s$ by \textit{a sequence of thirteen-dimensional feature vectors}. 
    \textbf{(2) The action space $\mathcal{A}$.}  
    To take into account the ratio and order of selected cuts, we define the action space by \textit{all the ordered subsets} of the candidate cuts $\mathcal{C}$. 
    It can be challenging to explore the action space efficiently, as the cardinality of the action space can be extremely large due to its combinatorial structure.
    \textbf{(3) The reward function $r$.} 
    To evaluate the impact of the added cuts on solving MILPs, we design the reward function by (i) measures collected at the end of solving LP relaxations such as the dual bound improvement, (ii) or end-of-run statistics, such as the solving time and the primal-dual gap integral. For the first, the reward $r(s,a)$ can be defined as the negative dual bound improvement at each step. For the second, the reward $r(s,a)$ can be defined as zero except for the last step $(s_T,a_T)$ in a trajectory, i.e., $r(s_T,a_T)$ is defined by the negative solving time or the negative primal-dual gap integral. 
    \textbf{(4) The transition function $f$.} The transition function maps the current state $s$ and the action $a$ to the next state $s^{\prime}$, where $s^{\prime}$ represents the next LP relaxation generated by adding the selected cuts at the current LP relaxation. 
    \textbf{(5) The terminal state.}
    There is no standard and unified criterion to determine when to terminate the cut separation procedure \citep{l2c_lookahead}. Suppose we set the cut separation rounds as $T$, then the solver environment terminates the cut separation after $T$ rounds.
    \item \textbf{Hierarchical Sequence Model} The policy network architecture of HEM is also illustrated in Figure \ref{fig:hem/illustration}. \textbf{First}, the higher-level policy learns the number of cuts that should be selected by predicting a proper ratio. Suppose the length of the state is $N$ and the predicted ratio is $k$, then the predicted number of cuts that should be selected is $\lfloor N*k \rfloor$, where $\lfloor \cdot \rfloor$ denotes the floor function. We define the higher-level policy by $\pi^h: \mathcal{S} \to \mathcal{P}(\left[0,1\right])$, where  $\pi^h(\cdot|s)$ denotes the probability distribution over $\left[0,1\right]$ given the state $s$.  \textbf{Second}, the lower-level policy learns to select an ordered subset with the size determined by the higher-level policy.
    We define the lower-level policy by $\pi^l: \mathcal{S} \times \left[0,1\right] \to \mathcal{P}(\mathcal{A})$, where $\pi^l(\cdot|s,k)$ denotes the probability distribution over the action space given the state $s$ and the ratio $k$. Specifically, 
    we formulate the lower-level policy as a sequence model, which can capture the interaction among cuts. \textbf{Finally}, we derive the cut selection policy via the law of total probability, i.e.,
    $\pi(a_k|s) = \mathbb{E}_{k\sim \pi^h(\cdot|s)}[\pi^l(a_k|s,k)],$
    where $k$ denotes the given ratio and $a_k$ denotes the action. The policy is computed by an expectation, as $a_k$ cannot determine the ratio $k$. For example, suppose that $N=100$ and the length of $a_k$ is $10$, then the ratio $k$ can be any number in the interval $[0.1,0.11)$. Actually, we sample an action from the policy $\pi$ by first sampling a ratio $k$ from $\pi^h$ and then sampling an action from $\pi^l$ given the ratio.
    \item \textbf{Training: Hierarchical Policy Gradient} For the cut selection task, we aim to find $\theta$ that maximizes the expected reward over all trajectories
    \begin{align}\label{eq:obj}
        J(\theta) = \mathbb{E}_{s\sim \mu, a_k \sim \pi_{\theta}(\cdot|s)}[r(s,a_k)],
    \end{align}
    where $\theta=
    \left[\theta_1,\theta_2\right]$ with $\left[\theta_1,\theta_2\right]$ denoting the concatenation of the two vectors, $\pi_{\theta}(a_k|s) = \mathbb{E}_{k\sim \pi^h_{\theta_1}(\cdot|s)}[\pi^l_{\theta_2}(a_k|s,k)]$, and $\mu$ denotes the initial state distribution. To train the policy with a hierarchical structure, we derive a hierarchical policy gradient.
    \begin{proposition}\label{proof_hpg}
        Given the cut selection policy $\pi_{\theta}(a_k|s) = \mathbb{E}_{k\sim \pi^h_{\theta_1}(\cdot|s)}[\pi^l_{\theta_2}(a_k|s,k)]$ and the training objective (\ref{eq:obj}), the hierarchical policy gradient takes the form of 
        \begin{align*}
            \nabla_{\theta_1}J(\left[\theta_1,\theta_2\right])
            & = \mathbb{E}_{s\sim\mu, k\sim \pi^h_{\theta_1}(\cdot|s)} [\nabla_{\theta_1} \log(\pi^h_{\theta_1}(k|s)) \mathbb{E}_{a_k\sim \pi^l_{\theta2}(\cdot|s,k)}[r(s,a_k)] ], \\
            \nabla_{\theta_2}J(\left[\theta_1,\theta_2\right])
            & = \mathbb{E}_{s\sim\mu, k\sim \pi^h_{\theta_1}(\cdot|s), a_k\sim \pi^l_{\theta_2}(\cdot|s,k)} [\nabla_{\theta_2}\log \pi^l_{\theta_2}(a_k|s,k) r(s,a_k)].
        \end{align*}
    \end{proposition}
    We use the derived hierarchical policy gradient to update the parameters of the higher-level and lower-level policies. We implement the training algorithm in a parallel manner that is closely related to the asynchronous advantage actor-critic (A3C) \citep{a3c}. 
\end{enumerate}

\begin{table*}[t]
\caption{Evaluation of HEM on real-world Production Planning problems.}
    \centering
    \resizebox{0.8\textwidth}{0.08\textheight}{
        \begin{tabular}{@{}ccccc@{}}
        \toprule
         & \multicolumn{4}{c}{Production Planning ($n=3582.25,\,\,m=5040.42$)} \\ \midrule
        \multirow{2}{*}{Method} & \multirow{2}{*}{Time (s) $\downarrow$} & \multirow{2}{*}{\begin{tabular}[c]{@{}c@{}}Improvement $\uparrow$ \\      (Time, \%)\end{tabular}} & \multirow{2}{*}{PD integral $\downarrow$} & \multirow{2}{*}{\begin{tabular}[c]{@{}c@{}}Improvement $\uparrow$\\      (PD integral, \%)\end{tabular}} \\
         &  &  &  &  \\ \midrule
        NoCuts & 278.79 (231.02) & NA & 17866.01 (21309.85) & NA \\
        Default & 296.12 (246.25) & -6.22 & 17703.39 (21330.40) & 0.91 \\
        Random & 280.18 (237.09) & -0.50 & 18120.21 (21660.01) & -1.42 \\
        NV & 259.48 (227.81) & 6.93 & 17295.18 (21860.07) & 3.20 \\
        Eff & 263.60 (229.24) & 5.45 & 16636.52 (21322.89) & 6.88 \\ \midrule
        SBP & 276.61 (235.84) & 0.78 & 16952.85 (21386.07) & 5.11 \\
        HEM (Ours) & \textbf{241.77 (229.97)} & \textbf{13.28} & \textbf{15751.08 (20683.53)} & \textbf{11.84} \\ \bottomrule
        \end{tabular}
    }
\label{tab:hem/evaluation}
\end{table*}

Please refer to \cite{hem_iclr} for more details. 

\textbf{Deployment and Performance Gain} 
We deploy HEM to large-scale real-world production planning problems from Huawei.
The results in Table \ref{tab:hem/evaluation} show that HEM significantly outperforms all the baselines in terms of the Time and PD integral. 
The results demonstrate the strong ability to enhance the Optverse solver with our proposed HEM in real-world applications.

\subsubsection{Neural Heuristics}
\paragraph{Background} Finding a good feasible solution in a short time is a very challenging task in solving Mixed Integer Linear Programming (MILP) problems. One of the most promising methods for leveraging machine learning to solve it is a Neural Diving heuristic \cite{neuraldiving}.
Without loss of generality, we will restrict the description only to the binary MILPs.
The main idea behind the Neural Diving method is to learn to generate correct partial assignments to selected variables of a given problem in order to solve corresponding sub-problems.
Providing that Neural Diving correctly predicts these assignments the problem could be solved much faster.

\paragraph{Problem Setup} A deep neural network is used to produce multiple {\it partial
assignments} of the binary variables of the original MILP problem.
The remaining unassigned variables define smaller ``sub-problems'' (sub-MILPs), which are solved using an off-the-shelf solver to obtain solutions. 
This method can be easily parallelized since every sub-MILP can be solved independently.  
Neural Diving can indeed help to quickly get good solutions, though it does not guarantee global optimality.

\paragraph{Method}
Following~\cite{Exact}, we encoded each MILP as a bi-partite graph with two sets of vertices, representing its variables and constraints, and gave it as input to a Graph Convolutional Neural Network (GCNN) that can predict the values of binary variables.
Such a network can be trained in a supervised manner using the binary cross-entropy loss.
We enhanced GCNN architecture with layer norms and skip connections to make the training more stable, utilized the primal integral~\cite{berthold} for performance evaluation.
We observed that performance of the neural diving method is mostly influenced by the following key factors: (1)~by the number of parallel sub-MILPs, (2)~by their sizes which are determined by the size of partial assignments, and, what is also very important, (3)~by the choice of binary variables that constitute these partial assignments.
While the first factor is only related to the available hardware resources and the second can be balanced by fine-tuning, the optimal choice of the variables is not so straight forward.

\begin{figure}[t]
\centering
\includegraphics[scale=0.60]{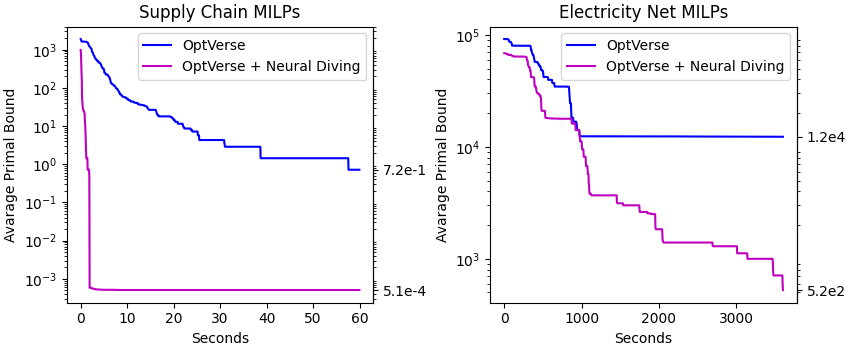}
\caption{Performance of the neural diving method on two real-life problems}
\label{fig:experiments}
\end{figure}

\paragraph{Deployment and Performance Gain} To make a proper choice of the variables, we must take into account the confidence of their prediction and, at the same time, also consider the CPU time which is spent on solving corresponding sub-MILPs.
To tackle this trade-off, we developed our own method that is based on ranking the variables according to their ability to reduce the CPU time that would be required to solve the MILP if they were assigned.
We proved that \textbf{good rankings} can be learnt off-line and used during the inference to generate faster sub-MILPs.
Also we enhanced the GCNN training procedure by using auxiliary losses, one for pumping the feasibility of generated sub-MILPs and the other for better balance of 0 and 1 values in partial assignments. We performed tests on various synthetic and real life datasets which proved that it can provide 3 to 10x speedup compared to the baseline version of OptVerse AI solver (see Figure~\ref{fig:experiments}).

\subsection{Parameter Tuning Framework}
To help users and researchers thoroughly explore hyper-parameter space of OptVerse AI solver, we developed a Parameter Tuning Framework integrating various tuners as a tool to improve searching efficiency and the convenience of tuning operation. Tuner generates configuration - a set of parameters based on search space with further processing by solver. Observed result used to improve generation of new parameters in the next step. The whole process is repeated until maximum amount of trials is reached or other stopping criteria is met. 

\subsubsection{Overview}

There are several main concepts used to describe high-level functionality of our developed framework.
\begin{itemize}
\item Experiment: single task of finding optimal hyper-parameters. It consist of several trials;
\item Search space: list of hyper-parameters with corresponding names, types and ranges;
\item Configuration: set of hyper-parameters instantiated from search space
\item Trial: execution of configuration on solver using available hardware (HW)
\end{itemize}
To start new experiment user needs to supply search space file, define hardware to execute trials. If it is necessary to change logic of solver execution or use different solver it might be needed to adjust code of executed model.

\begin{figure}[t]
\centering
  \includegraphics[width=\linewidth]{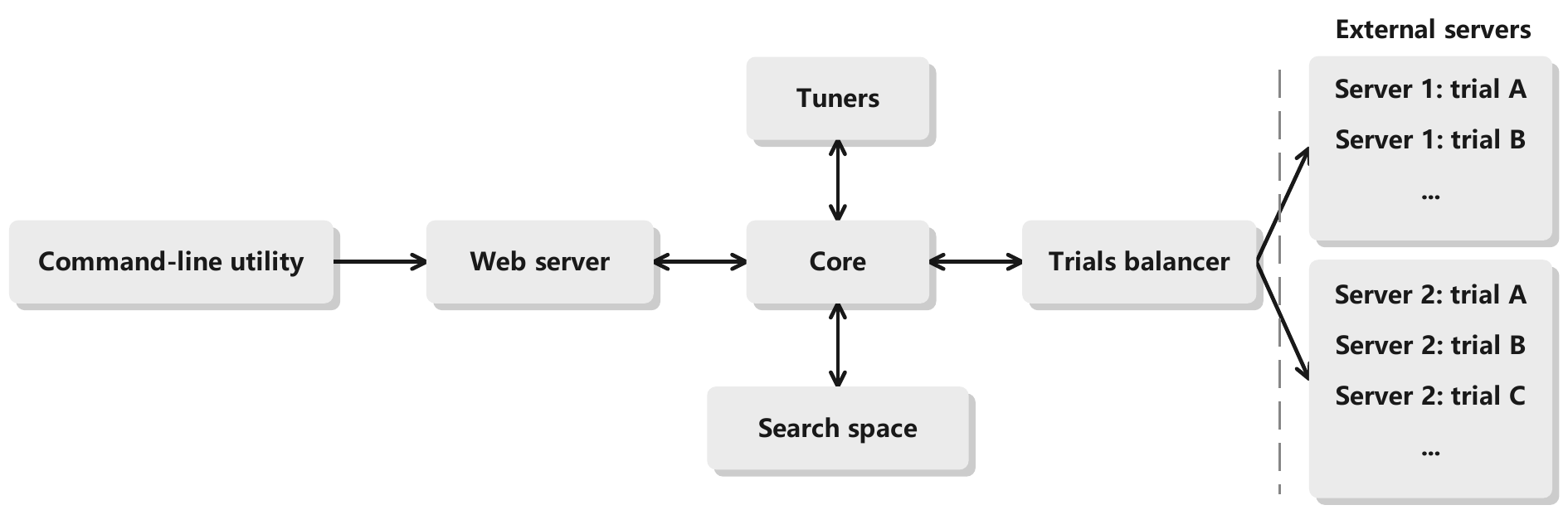}
  \caption{High-level architecture of our proposed hyper-parameter tuning framework for OptVerse AI solver, which are featured with (1) command-line utility used to start new experiment, (2) monitor general status of executed experiments or to stop active one, and (3) web server has rich UI functionality to display details and statistics of experiment execution process.}
  \label{fig:arch1}
\end{figure}

Figure \ref{fig:arch1} shows high-level architecture of our proposed hyper-parameter tuning framework for OptVerse AI solver. Command-line utility used to start new experiment, monitor general status of executed experiments or to stop active one. It is necessary to prepare search space and experiment configuration including hardware mapping in separate configuration files to start new experiment. Web server of the framework has rich UI functionality to display details and statistics of experiment execution process together with best available solution. At the same time web server has additional endpoints to collect statistics and status of running trials, supported from remote servers.
To improve stability of whole solution algorithms of a tuner is launched in a separate process. Core block of the framework is responsible for trial`s execution orchestration and preparation of list of a tasks for trials balancer, which is equipped with our proposed hyper-parameter tuning algorithm, namely HEBO~\cite{hebo}, Transformber BO~\cite{alex2023end} and variant of Differential Evolution (DE)~\cite{Storn1997}. Trials balancer distributes execution of a trials among available and mapped hardware. It is possible to run several trials within same server if hardware requirements are fulfilled. It is also possible to run trials in data-parallel way using several servers.

\subsubsection{Parameter Tuning Algorithms}

\paragraph{HEBO \& Transformer BO} Since it is hard for practitioners to explicitly model the relationship between the solver's performance and corresponding hyper-parameters, performance function of hyper-parameters is actually a black-box function. Hence, Bayesian Optimization (BO), which is a classic black-box optimization technique with high sample efficiency, is widely-used for tuning solvers. 

Despite that the BO framework shows availability in tuning tasks for solvers, there still exist some issues in practice~\cite{hebo}. Firstly, Gaussian Process (GP), which is a common surrogate model in BO, typically requires two important modelling assumptions stated as data stationarity and homoscedasticity of the noise distribution. Unfortunately, even simple hyper-parameter tuning tasks can exhibit significant non-stationarity and heteroscedastic. Consequently, a distinct approach is required to process such non-ideal data in real scenario. Secondly, there are various acquisition functions trading off exploration and exploitation to the approximated surrogate model in general BO framework. However, various acquisition functions may result in opposing solutions when pursuing their own optima. Thus, we also need to tackle the conflicting acquisitions.

To address the aforementioned issues, we integrate Heteroscedastic and Evolutionary Bayesian Optimization (HEBO) algorithm which is also developed by ourselves in our parameter tuning framework~\cite{hebo}. This algorithm introduces an Kumaraswamy input warping which performs a transformation to the input data (i.e., hyper-parameter variables) and corrects the non-stationarity. Furthermore, to tackle the data heteroscedasticity, HEBO leverage ideas from warped GP~\cite{warpedGP} where the Box-Cox transformation is considered as a corrective mapping for non-Gaussian data. As for the conflicting acquisition functions, HEBO designs a multi-objective acquisition function that combines three different commonly-used acquisitions, i.e., Expected Improvement (EI), Probability of Improvement (PI), and Upper Confidence Bound (UCB). Formally, we solve
\begin{equation}
\max \limits_{\mathbf{x}} \left(\bar{\alpha}^{\theta}_{EI}(\mathbf{x}|\mathcal{D}), \bar{\alpha}^{\theta}_{PI}(\mathbf{x}|\mathcal{D}), \bar{\alpha}^{\theta}_{UCB}(\mathbf{x}|\mathcal{D})\right)
\end{equation}
where $\bar{\alpha}^{\theta}_{(\cdot)}(\mathbf{x}|\mathcal{D})$ denotes a single acquisition in which $\mathcal{D}$ and $\mathbf{x}$ are historical data (hyper-parameter configurations and corresponding solver performance) and newly sampled points. This comprehensive formulation avoids the hard selection of an optimal acquisition such that the solution will not be dominated by one single acquisition. Correspondingly, HEBO employs the non-dominated sorting genetic algorithm II (NSGA-II) for seeking Pareto-front solutions of the multi-objective acquisitions. With all designs mentioned above, HEBO significantly outperforms existing black-box optimisers on 108 machine learning hyperparameter tuning tasks comprising the Bayesmark benchmark, and win the championship of the NIPS 2020 Black-Box Optimisation challenge~\cite{hebo}.

To further improve sample efficiency on new-coming \textit{target} tasks, one available approach is to use meta BO to transfer knowledge from related \textit{source} tasks in between~\cite{bai2023transfer}. We also propose an end-to-end meta BO framework termed \textit{Transformer BO}~\cite{alex2023end} for solver tuning. This framework uses a new transformer-based neural process to directly predict acquisition function values from observed data, without relying on traditional surrogate models like GP. After collecting enough observed data from \textit{source} tasks, an improved reinforcement learning protocol with inductive bias is adopted to train the Transformer architecture, and then the well-trained meta neural process can be used for tuning \textit{target} tasks more efficiently with prior knowledge from \textit{source} tasks.

\paragraph{Differential evolution and its modifications}

Differential Evolution (DE)~\cite{Storn1997} is a powerful and versatile optimization algorithm that falls under the category of evolutionary algorithms, which are inspired by the process of natural selection and evolution. DE operates on a population of candidate solutions and iteratively refines them to find the optimal solution. It uses mutation, crossover, and selection operations to update the population. The main steps of the DE algorithm are as follows: \textbf{(1) Initialization}: Initialize a population of candidate solutions (vectors) randomly within predefined bounds. \textbf{(2) Mutation}: Create new candidate solutions by perturbing the current population using mutation strategies. In particular, the mutant vector $\mathbf{v}$ may be taken as $\mathbf{v} = \mathbf{a} + F (\mathbf{b} - \mathbf{c})$, where $\mathbf{a}$, $\mathbf{b}$ and $\mathbf{c}$ are three random distinct vectors from the population. \textbf{(3) Crossover}: Change the coordinates of the mutated solution $\mathbf{v}$ by corresponding coordinates of the current vector $\mathbf{x}$ with the cross-over probability $C$ to create trial solutions. \textbf{(4) Selection}: Evaluate the objective function for the trial vector and select the candidates that produce better results, replacing the current population with the best candidates. \textbf{(5) Termination Criteria}: Determine when to stop the algorithm, usually based on a maximum number of iterations, a convergence threshold or the value of the fitness function to be optimized.

DE is known for its ability to efficiently explore the search space and converge to optimal solutions when tackling problems with non-linearity, high-dimensionality, and multimodality. Throughout various competitions organized under the IEEE Congress on Evolutionary Computation conference series, different variants of DE consistently achieve top rankings \cite{Das2016}.

The efficiency of the classical implementation of DE depends on three crucial control parameters, namely, the scale factor, crossover rate, and population size. In particular, the scale factor and crossover rate play a pivotal role in determining the algorithm's behaviour, enabling a fine balance between global and local search, often referred to as exploration and exploitation \cite{repinek2013}. A larger population size introduces more diversity into the population and tend to perform better in global search, as they have a higher chance of covering various areas in the search space, although at higher computational cost. Through adjusting the controlling strategies of aforementioned three parameters, some variants of DE can be acquired, which are summarized in Table \ref{tab:DEmods}.

\begin{table}[t]
\caption{Modifications of Differentiable Evolutionary}

\centering
    \begin{tabular}{cp{6cm}c}
\toprule
         Algorithm & Description & Reference \\
\hline
         JADE & JADE adapts the mutation and crossover strategies during the optimization process, making it more efficient by dynamically controlling the parameters. & \cite{JingqiaoZhang2009}\\
\hline
         SHADE & SHADE is an adaptive DE which uses a historical memory of successful control parameter settings to guide the selection of future control parameter values. & \cite{SHADE}\\
\hline
         L-SHADE & L-SHADE extends SHADE with Linear Population Size Reduction, which continually decreases the population size according to a linear function. & \cite{LSHADE}\\
\bottomrule
         
    \end{tabular}
    \label{tab:DEmods}
\end{table}

To analyze the efficiency of algorithms, we choose a pool of testing functions. They include:
\begin{itemize}
	\item convex quadratic functions with different large number of dimensions;
	\item polynomial curve fitting task;
	\item non-convex functions: Rosenbrock's, Rastrigin's, Ackley's, Griewank's, Schaffer's F6, HGBat, Schwefel, Weierstrass.
\end{itemize}

Based on our numerical tests for chosen pool of test problems we propose an advanced version of DE which we refer to LJADE. Its features are summarized below.

We use  rand-to-p-best1 strategy, described in \cite{Zhang2009}, that is expressed as follows:
\begin{equation}
    \textbf{rand-to-p-best/1: } \mathbf{v} = \mathbf{a} + F (\mathbf{g_p} - \mathbf{a} ) +  F (\mathbf{b} - \mathbf{c} ),
\end{equation}
where $\mathbf{g_p}$ is the random vector from a pool of the top $\mathbf{p}$ individuals in the current population  with $\mathbf{p}\in (0,1]$ * 100 \%. We also use the self-adaptation scheme for the mutation factor~\cite{JingqiaoZhang2007} to independently select parameters that favour the survival of individuals in case they are not initialized optimally enough, and introduce the modified scheme for the crossover probability~\cite{FeiPeng2009} to balance the bias towards small values during self-adaptation process. Furthermore, the population size is dynamically reduced using a linear population size reduction method~\cite{LSHADE} and an initialization scheme based on quasi-random Halton sequences~\cite{Halton1964} is leveraged, which is somewhat superior to random initialization scheme.


The underlying concept behind these modifications is that the algorithm can adapt its approach based on the rate at which the residual function is decreasing. This adaptability allows it to switch between more aggressive searches and finer local searches as needed.

\paragraph{Deployment and Performance Gain}
The results of experiments with several instances from MIPLIB2017 benchmark are shown in Table \ref{scipresults}. To optimize hyper-parameters, 1000 trials of tuning are used,
while the NNI concurrency is 25. The population size of LJADE is set to 75.

\begin{table}[t]
\caption{Tuning results on MIPLIB2017 benchmark with SCIP solver}
\centering
 \begin{tabular}{p{0.13\textwidth}p{0.2\textwidth}p{0.1\textwidth}p{0.2\textwidth}p{0.1\textwidth}} 
 \toprule
Instance & Default solving time &Tool &Optimized solving time, speedup &Tuning time \\
\hline
\multirow{2}{*}{app1-1.mps} &	\multirow{2}{*}{3.34s} & SMAC & 0.70s, 4.77x & 21 m	\\
		& & LJADE	& 0.87s, 3.84x & 22m \\
\hline
\multirow{2}{*}{air05.mps} &\multirow{2}{*}{ 25.03s} & SMAC	& 28.69, 0.87x	& 8h 37m\\
            & &LJADE & 20.0s, 1.25x	& 49m \\
\hline
\multirow{2}{*}{neos8.mps} & \multirow{2}{*}{3.69s } & SMAC	 & 3.41s, 1.08x& 1h 4m \\
	& &	LJADE & 4.02s, 0.9x & 23m 23s \\
\hline
\multirow{2}{*}{nu25-pr12.mps} &\multirow{2}{*}{ 3.12s }& SMAC	 & 1.91s, 1.63x & 37 m\\	
	& &	LJADE & 1.46s, 2.14x & 23m \\
\hline
\multirow{2}{*}{eil33-2.mps} &\multirow{2}{*}{ 50.68s }& SMAC & 29.30s, 1.73x & 8h 30m\\		
	& &	LJADE & 35.42s, 1.43x & 1h 12m \\
\hline
\multirow{2}{*}{swath3.mps} &\multirow{2}{*}{ 916.66s} & SMAC & 41.85s, 21.9x& 21h 28m\\		
      & & LJADE	& 74.31s, 12.33x & 6h 58m\\
 \bottomrule
 \end{tabular}
\label{scipresults}
\end{table}

\section{Conclusion and Outlook}

In conclusion, our investigation into the OptVerse AI Solver reveals a monumental leap in digital construction capacities across various industries. By harmoniously weaving machine learning (ML) with operational research, we have endowed the solver with unparalleled adaptation and efficiency. The induction of sophisticated ML techniques, such as data generation, policy learning, and hyper-parameter tuning, has broadened the scope of solvable mathematical programming instances, enabling solvers to surmount the confines of traditional data scarcity and proprietary constraints. Empirical advances in convergence rates and solver performance, underscored by our substantial augmentation in speed and precision over established benchmarks and real-world problems, affirm the transformative potential of ML within this domain.

Looking ahead, the synergy between the OptVerse AI Solver and cutting-edge large language models (LLMs) opens up expansive prospects for further refinement and application. LLMs could revolutionize human-solver interaction, offering intuitive natural language interfaces for problem formulation and solution interpretation, thus democratizing accessibility. They could also contribute to the automated generation of problem instances and facilitate the extraction of nuanced patterns in data, enriching the solver's learning landscape. This confluence of traditional machine learning techniques with large-scale language processing promises a future where solvers not only possess heightened computational acumen but also exhibit a near-cognitive understanding of complex challenges, heralding a new frontier in the realm of combinatorial optimization.

\section*{Acknowledgements}
The authors would like to thank all the participants and developers of OptVerse AI Solver, including Huawei Vancouver Research Center, Huawei Moscow Research Center, Huawei Minsk Research Center, and Huawei Munich Research Center. This work was fully supported by Huawei Cloud Computing Technologies Co., Ltd.

\bibliographystyle{unsrtnat}
\bibliography{library}


\end{document}